\documentclass{article}

\usepackage[labelfont=it,labelsep=period,font=small,skip=0pt,belowskip=-0.75em]{caption}

\usepackage{microtype}
\usepackage{graphicx}
\usepackage{subfig}
\usepackage{booktabs} 
\usepackage[T1]{fontenc}
\usepackage{authblk}
\usepackage{natbib}

\usepackage{hyperref}

\usepackage[accepted]{icml2022-arxiv}

\makeatletter
\renewcommand\AB@affilsepx{\qquad \protect\Affilfont}
\makeatother

\usepackage{amsmath}
\usepackage{amssymb}
\usepackage{mathtools}
\usepackage{amsthm}

\usepackage[capitalize,noabbrev]{cleveref}

\theoremstyle{plain}

\theoremstyle{definition}

\newcommand{\E}{\mathop{\mathbb{E}}}      
\newcommand{\R}{\mathbb{R}}      
\newcommand{\innerproduct}[2]{\langle #1, #2 \rangle}
\newcommand{\projB}{{ {\footnotesize \textsf{proj}}_{ B}}}
\newcommand{\projBComp}{{ {\footnotesize \textsf{proj}}_{ B_{\perp}}}}

\newcommand{\hh}{\boldsymbol h} 
\newcommand{\uu}{\boldsymbol u} 
\newcommand{\vv}{\boldsymbol v} 
\newcommand{\xx}{\boldsymbol x} 
\newcommand{\yy}{\boldsymbol y} 

\makeatletter
\newcommand*\bigcdot{\mathpalette\bigcdot@{.5}}
\newcommand*\bigcdot@[2]{\mathbin{\vcenter{\hbox{\scalebox{#2}{$\m@th#1\bullet$}}}}}
\makeatother

\begin{document}

\twocolumn[
\icmltitle{Eliciting Latent Predictions from Transformers with the Tuned Lens}

\begin{icmlauthorlist}
\icmlauthor{Nora Belrose}{eai,far}
\icmlauthor{Igor Ostrovsky}{eai}
\icmlauthor{Lev McKinney}{toronto,far}
\icmlauthor{Zach Furman}{eai,bu}
\icmlauthor{Logan Smith}{eai}
\icmlauthor{Danny Halawi}{eai}
\icmlauthor{Stella Biderman}{eai}
\icmlauthor{Jacob Steinhardt}{ucb}
\end{icmlauthorlist}

\icmlaffiliation{eai}{Eleuther AI}
\icmlaffiliation{far}{FAR AI}
\icmlaffiliation{bu}{Boston University}
\icmlaffiliation{toronto}{University of Toronto}
\icmlaffiliation{ucb}{UC Berkeley}

\icmlcorrespondingauthor{Nora Belrose}{nora@eleuther.ai}

\vskip 0.3in
]

\date{\vspace{-4ex}}

\begin{abstract}

We analyze transformers from the perspective of iterative inference, seeking to understand how model predictions are refined layer by layer. To do so, we train an affine probe for each block in a frozen pretrained model, making it possible to decode every hidden state into a distribution over the vocabulary. Our method, the \emph{tuned lens}, is a refinement of the earlier ``logit lens'' technique, which yielded useful insights but is often brittle. 

We test our method on various autoregressive language models with up to 20B parameters, showing it to be more predictive, reliable and unbiased than the logit lens. With causal experiments, we show the tuned lens uses similar features to the model itself. We also find the trajectory of latent predictions can be used to detect malicious inputs with high accuracy. All code needed to reproduce our results can be found at \url{https://github.com/AlignmentResearch/tuned-lens}.
\end{abstract}

\section{Introduction}
\label{introduction}

\begin{figure}[th]
\hspace{-1em}
\includegraphics[scale=0.45, clip]{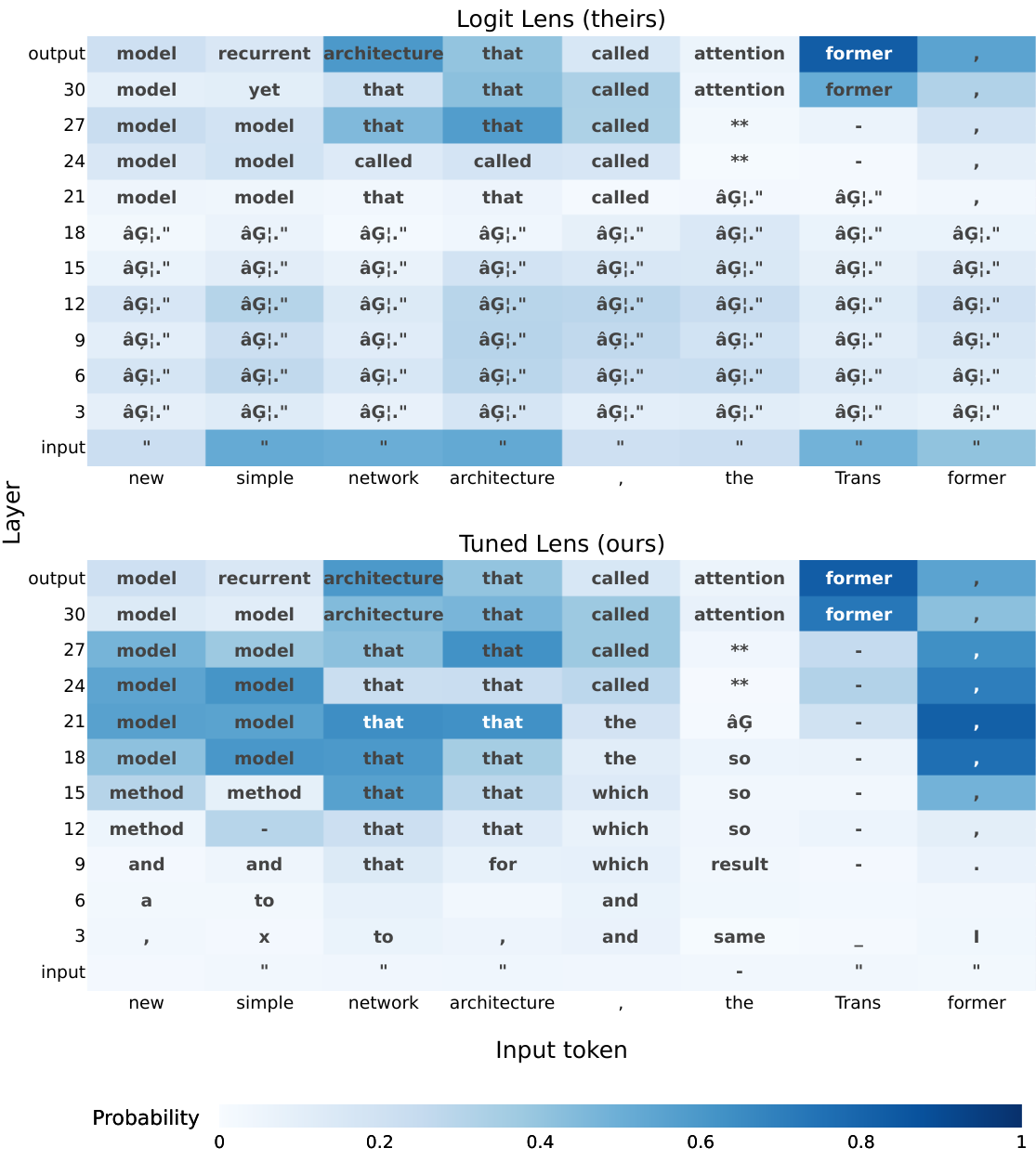}
\caption{Comparison of our method, the \emph{tuned lens} (bottom), with the ``logit lens'' (top) for GPT-Neo-2.7B prompted with an except from the abstract of \citet{vaswani2017attention}. Each cell shows the top-1 token predicted by the model at the given layer and token index. The logit lens fails to elicit interpretable predictions before layer 21, but our method succeeds.}
\label{fig:front-page}
\end{figure}

\printAffiliationsAndNotice{}
The impressive performance of transformers in natural language processing \cite{brown2020language} and computer vision \cite{dosovitskiy2020image} suggests that their internal representations have rich structure worthy of scientific investigation. One common approach is to train classifiers to extract specific concepts from hidden states, like part-of-speech and syntactic structure \cite{hewitt2019structural, tucker2021if, li2022emergent}.

In this work, we instead examine transformer representations from the perspective of \emph{iterative inference} \cite{jastrzkebski2017residual}. Specifically, we view each layer in a transformer language model as performing an incremental update to a latent prediction of the next token.\footnote{See Appendix \ref{app:iterative-inference} for evidence supporting this view, including novel empirical results of our own.} We decode these latent predictions through early exiting, converting the hidden state at each intermediate layer into a distribution over the vocabulary. This yields a sequence of distributions we call the \emph{prediction trajectory}, which exhibits a strong tendency to converge smoothly to the final output distribution, with each successive layer achieving lower perplexity.

We build on the ``logit lens'' \cite{nostalgebraist2020logitlens}, an early exiting technique that directly decodes hidden states into vocabulary space using the model's pretrained unembedding matrix. We find the logit lens to be unreliable (Section~\ref{logit-lens}), failing to elicit plausible predictions for models like BLOOM \cite{scao2022bloom} and GPT Neo \cite{black2021gpt}. Even when the logit lens appears to work, its outputs are hard to interpret due to \emph{representational drift}: features may be represented differently at different layers of the network. Other early exiting procedures also exist \cite{schuster2022confident}, but require modifying the training process, and so can't be used to analyze pretrained models.

To address the shortcomings of the logit lens, we introduce the \emph{tuned lens}. We train $L$ affine transformations, one for each layer of the network, with a distillation loss: transform the hidden state so that its image under the unembedding matches the final layer logits as closely as possible. We call these transformations \emph{translators} because they ``translate'' representations from the basis used at one layer of the network to the basis expected at the final layer. Composing a translator with the pretrained unembedding yields a probe \cite{alain2016understanding} that maps a hidden state to a distribution over the vocabulary.

We find that tuned lens predictions have substantially lower perplexity than logit lens predictions, and are more representative of the final layer distribution. We also show that the features most influential on the tuned lens output are also influential on the model itself (Section \ref{causal-fidelity}). To do so, we introduce a novel algorithm called \emph{causal basis extraction} (CBE) and use it to locate the directions in the residual stream with the highest influence on the tuned lens. We then ablate these directions in the corresponding model hidden states, and find that these features tend to be disproportionately influential on the model output.

We use the tuned lens to gain qualitative insight into the computational process of pretrained language models, by examining how their latent predictions evolve during a forward pass (\cref{fig:front-page}, \cref{app:qualitative})

Finally, we apply the tuned lens in several ways: we extend the results of \citet{halawi2023overthinking} to new models (Section \ref{overthinking}), we find that tuned lens prediction trajectories can be used to detect prompt injection attacks \cite{perez2022ignore} often with near-perfect accuracy (Section \ref{prompt-injection}), and find that data points which require many training steps to learn also tend to be classified in later layers (Section \ref{example-difficulty}).

\section{The Logit Lens}
\label{logit-lens}

The logit lens was introduced by \citet{nostalgebraist2020logitlens}, who found that when the hidden states at each layer of GPT-2 \citep{radford2019language} are decoded with the unembedding matrix, the resulting distributions converge roughly monotonically to the final answer. More recently it has been used by \citet{halawi2023overthinking} to understand how transformers process few-shot demonstrations, and by \citet{dar2022analyzing}, \citet{geva2022transformer}, and \citet{millidge2022svd} to directly interpret transformer weight matrices.

\textbf{The method.} Consider a pre-LayerNorm transformer\footnote{Both the logit lens and the tuned lens are designed primarily for the pre-LN architecture, which is more unambiguously iterative. Luckily pre-LN is by far more common than post-LN among state-of-the-art models. See \citet[Appendix C]{zhang2020accelerating} for more discussion.} $\mathcal{M}$. We'll decompose $\mathcal{M}$ into two ``halves,'' $\mathcal{M}_{\leq\ell}$ and $\mathcal{M}_{>\ell}$. The function $\mathcal{M}_{\leq\ell}$ consists of the layers of $\mathcal{M}$ up to and including layer $\ell$, and it maps the input space to hidden states. Conversely, the function $\mathcal{M}_{>\ell}$ consists of the layers of $\mathcal{M}$ after $\ell$, which map hidden states to logits.

The transformer layer at index $\ell$ updates the representation as follows:
\begin{equation}
\label{eq:residual_equation}
 \hh_{\ell+1} = \hh_{\ell} + F_{\ell}(\hh_{\ell}),
\end{equation} where $F_{\ell}$ is the residual output of layer $\ell$.
\begin{figure}
    \centering
    \includegraphics[scale=0.65]{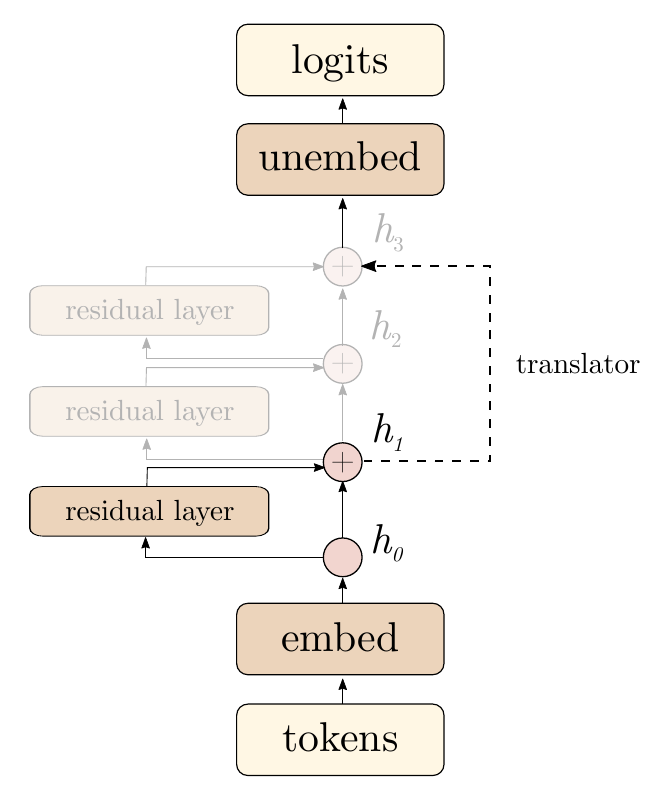}
    \caption{The tuned lens takes the hidden state at an intermediate layer (e.g. $h_1$ above), and applies a learned affine transformation (the translator). We then convert the hidden state into logits with the unembedding layer.}
    \label{fig:lens-diagram}
\end{figure}
Applying Equation \ref{eq:residual_equation} recursively, the output logits $\mathcal{M}_{>\ell}$ can be written as a function of an arbitrary hidden state $\hh_{\ell}$ at layer $\ell$: \begin{equation}
    \mathcal{M}_{>\ell}(\hh_{\ell}) = \mathrm{LayerNorm}\Big[\hh_{\ell} + \sum_{\ell'=\ell}^{L} \underbrace{F_{\ell'}(\hh_{\ell'})}_{\text{residual update}}\hspace{-0.08in}\Big]W_U.
    \label{eq:summed-residuals}
\end{equation}
The logit lens consists of setting the residuals to zero:
\begin{equation}
    \mathrm{LogitLens}(\hh_{\ell}) = \mathrm{LayerNorm}[\hh_{\ell}]W_U.
\end{equation}
While this simple form of the logit lens works reasonably well for GPT-2, \citet{nostalgebraist2021logitlens} found that it fails to extract useful information from the GPT-Neo family of models \citep{black2021gpt}. In an attempt to make the method work better for GPT-Neo, they introduce an extension which retains the last transformer layer, yielding: \begin{equation}
    \mathrm{LogitLens}^{\mathrm{ext}}(\hh_{\ell}) = \mathrm{LayerNorm}[\hh_{\ell} + F_L(\hh_{\ell})]W_U
    \label{eq:ll-extended}
\end{equation} This extension is only partially successful at recovering meaningful results; see Figure~\ref{fig:front-page} (top) for an example.

\textbf{Unreliability.}
Beyond GPT-Neo, the logit lens struggles to elicit predictions from several other models released since its introduction, such as BLOOM \citep{scao2022bloom} and OPT 125M \citep{zhang2022opt} (Figure~\ref{fig:opt-perplexity}).

Moreover, the type of information extracted by the logit lens varies both from model to model and from layer to layer, making it difficult to interpret. For example, we find that for BLOOM and OPT 125M, the top 1 prediction of the logit lens is often the \emph{input} token, rather than any plausible continuation token, in more than half the layers (\cref{fig:logit-lens-pathologies}).

\begin{figure}
\includegraphics[scale=0.5]{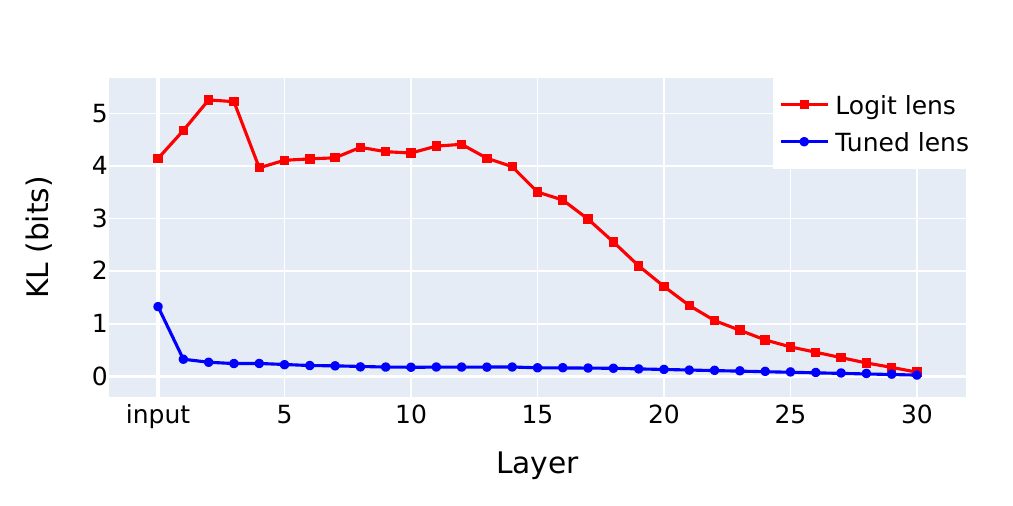}
\caption{Bias of logit lens and tuned lens outputs relative to the final layer output for GPT-Neo-2.7B. The last transformer layer is included for both probes. Unlike the tuned lens, the logit lens is systematically biased toward some vocabulary items over others until the very end of the network.}
\label{fig:bias-plot}
\end{figure}

\textbf{Bias.}
Even when the logit lens is useful, we find that it is a \emph{biased} estimator of the model's final output: it systematically puts more probability mass on certain vocabulary items than the final layer does.

This is concerning because it suggests we can't interpret the logit lens prediction trajectory as a belief updating in response to new evidence. The beliefs of a rational agent should not update in an easily predictable direction over time \cite{yudkowsky2007conservation}, since predictable updates can be exploited via Dutch books \cite{ramsey1926truth, de1937foresight, hacking1967slightly, garrabrant2016logical}. Biased logit lens outputs are trivially exploitable once the direction of bias is known: one could simply ``bet'' against the logit lens at layer $\ell < L$ that the next token will be one of the tokens that it systematically downweights, and make unbounded profit in expectation.

Let $\xx$ be a sequence of tokens sampled from a dataset $D$, and let $\xx_{<t}$ refer to the tokens preceding position $t$ in the sequence. Let $q_{\ell}(\cdot|\xx_{<t})$ be the logit lens distribution at layer $\ell$ for position $t$, and let $p(\cdot|\xx_{<t})$ be the final layer distribution for position $t$.

We define $p(v|\xx)$ to be the probability assigned to a vocabulary item $v$ in a sequence $\xx$, averaged over all positions $1 \ldots T$:
\begin{equation}
p(v|\xx) \overset{\mathrm{def}}{=} \frac{1}{T} \sum_{t=1}^{T} p(v|\xx_{<t}).
\end{equation}
Slightly abusing terminology, we say that $q_{\ell}$ is an ``unbiased estimator'' of $p$ if, for every item $v$ in the vocabulary, the probability assigned to $v$ averaged across all tokens in the dataset is the same:
\begin{equation}\label{eq:marginal-bias}
\begin{aligned}
    \E_{\xx \in D}\Big[q_{\ell}(v|\xx)\Big] &= \E_{\xx \in D}\Big[p(v|\xx)\Big]\\
    q_{\ell}(v) &= p(v)\\
    \forall v \in \mathcal{V},\quad \mathcal{V} &= \{\texttt{"aardvark"}, \ldots\}
\end{aligned}
\end{equation}
In practice, Equation \ref{eq:marginal-bias} will never hold exactly. We measure the degree of bias using the KL divergence between the marginal distributions, $D_{KL}(p\:||\:q_{\ell})$. 

In Figure \ref{fig:bias-plot} we evaluate the bias for each layer of GPT-Neo-2.7B. We find the bias of the logit lens can be quite large: around 4 to 5 bits for most layers. As a point of comparison, the bias of Pythia 160M's final layer distribution relative to that of its larger cousin, Pythia 12B, is just 0.0068 bits.

\section{The Tuned Lens}
\label{method}

\begin{figure}
    \centering
    \includegraphics[scale=0.5]{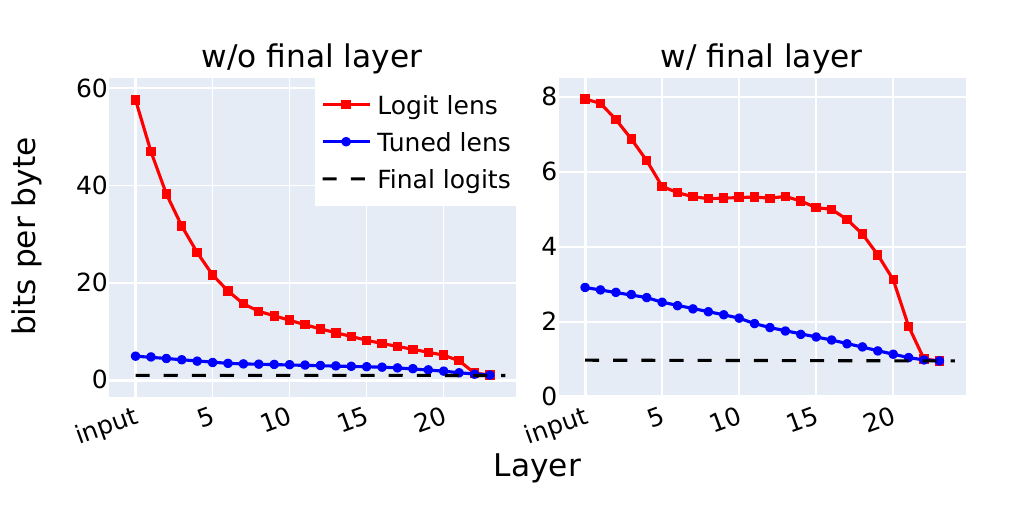}
    \caption{Perplexity of predictions elicited from BLOOM 560M under four conditions: the logit lens (red squares) and the tuned lens (blue circles), and including (left) and excluding (right) the final transformer layer from the probe. We find that tuned lens predictions have substantially lower perplexity whether or not the final layer is included, showing it is an independent and complementary proposal.}
    \label{fig:perplexity-bloom}
\end{figure}

\begin{figure*}[t]
    \centering
    \includegraphics[scale=0.5, clip]{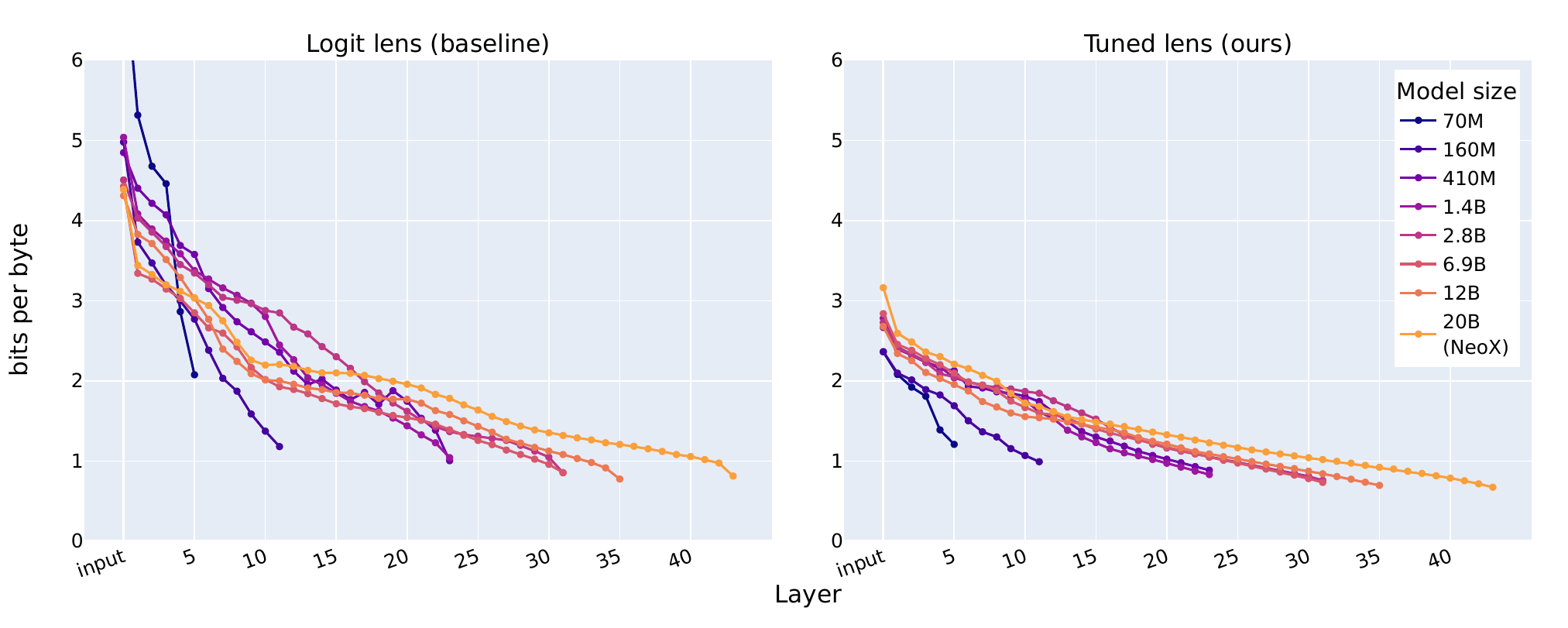}
    \caption{Perplexity of latent predictions elicited by the logit lens (left) and the tuned lens (right) from Pythia and GPT-NeoX-20B, as a function of layer index and model size. Tuned lens predictions are uniformly lower perplexity and exhibit lower variance across independently trained models.}
    \label{fig:pythia-perplexity}
\end{figure*}

One problem with the logit lens is that, if transformer layers learn to output residuals that are far from zero \emph{on average}, the input to $\mathrm{LogitLens}$ may be out-of-distribution and yield nonsensical results. In other words, the choice of zero as a replacement value is somewhat arbitrary-- the network might learn to rely on $\sum_{\ell'=\ell}^{L} \E[F_{\ell}(\hh_{\ell})]$ as a bias term.

Our first change to the method is to replace the summed residuals with a learnable constant value $\mathbf{b}_{\ell}$ instead of zero:
\begin{equation}
    \mathrm{LogitLens}^{\mathrm{debiased}}_{\ell}(\hh_{\ell}) = \mathrm{LogitLens}(\hh_{\ell} + \mathbf{b}_{\ell})
\end{equation}

\begin{figure}[t]
\centering
\includegraphics[clip, scale=0.5]{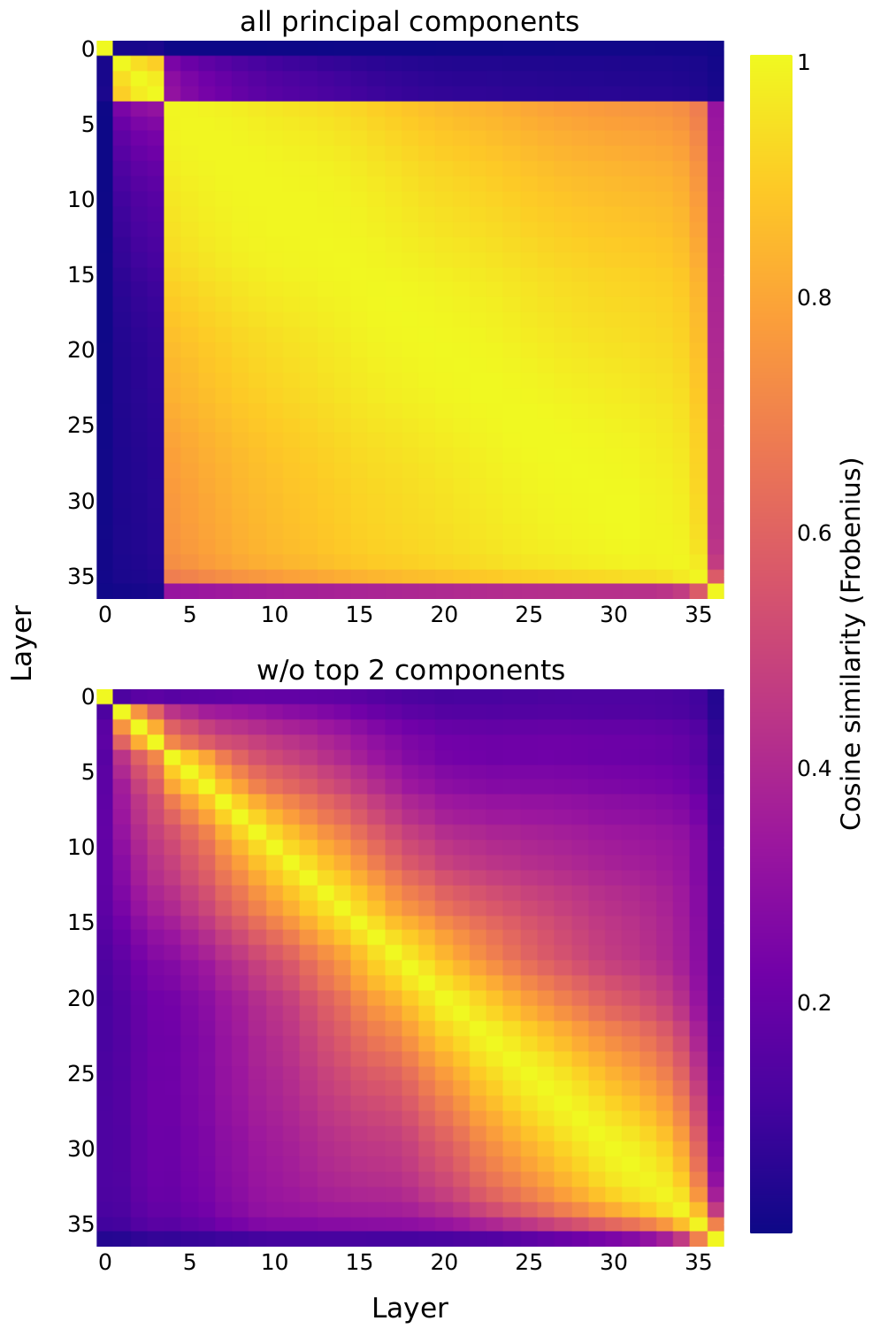}
\caption{Pairwise similarities of hidden state covariance matrices across layers of Pythia 12B. Layer 4 introduces two outlier dimensions which dominate the covariance; removing them reveals smooth representational drift with depth. To control for varying hidden state norms, we measure the Frobenius cosine similarity, or $\frac{\langle A, B \rangle_F}{ \|A\|_F \|B\|_F}$ for two matrices $A$ and $B$.}
\label{fig:covariance-drift}
\end{figure}

\textbf{Representation drift.} Another issue with the logit lens is that transformer hidden states often contain a small number of very high variance dimensions, and these ``rogue dimensions'' \citep{timkey2021all} tend to be distributed unevenly across layers; see Figure \ref{fig:covariance-drift} (top) for an example. Ablating an outlier direction can drastically harm performance \citep{kovaleva2021bert}, so if $\mathrm{LogitLens}$ relies on the presence or absence of particular outlier dimensions, the perplexity of logit lens predictions might be spuriously high.

Even when controlling for rogue dimensions, we observe a strong tendency for the covariance matrices of hidden states at different layers to drift apart as the number of layers separating them increases (Figure \ref{fig:covariance-drift}, bottom). The covariance at the final layer often changes sharply relative to previous layers, suggesting the logit lens might ``misinterpret'' earlier representations.

One simple, general way to correct for drifting covariance is to introduce a learnable change of basis matrix $A_{\ell}$, which learns to map from the output space of layer $\ell$ to the input space of the final layer. We have now arrived at the \emph{tuned lens} formula, featuring a learned affine transformation for each layer:
\begin{equation}
    \mathrm{TunedLens}_{\ell}(\hh_{\ell}) = \mathrm{LogitLens}(A_{\ell}\hh_{\ell} + \mathbf{b}_{\ell})
\end{equation}
We refer to $(A_{\ell}, \mathbf{b}_{\ell})$ as the \emph{translator} for layer $\ell$.

\textbf{Loss function.}
We train the translators to minimize KL between the tuned lens logits and the final layer logits:
\begin{equation}
    \mathop{\mathrm{argmin\:}} \E_{\xx}\Big[D_{KL}(f_{>\ell}(\hh_{\ell})\,||\,\mathrm{TunedLens}_{k}(\hh_{\ell}))\Big]
\end{equation}
where $f_{>\ell}(\hh_{\ell})$ refers to the rest of the transformer after layer $\ell$. This can be viewed as a distillation loss, using the final layer distribution as a soft label \citep{sanh2019distilbert}. It ensures that the probes are not incentivized to learn extra information over and above what the model has learned, which can become a problem when training probes with ground truth labels \citep{hewitt2019designing}.

\textbf{Implementation details.} When readily available, we train translators on a slice of the validation set used during pretraining, and use a separate slice for evaluation. Since BLOOM and GPT-2 do not have publicly available validation sets, we use the Pile validation set \citep{gao2020pile,biderman2022datasheet}. The OPT validation set is also not publicly available, but a member of the OPT team helped us train a tuned lens on the OPT validation set. Documents are concatenated and split into uniform chunks of length 2048.

We evaluate all models on a random sample of 16.4M tokens from their respective pretraining validation sets. We leave out the final transformer layer for GPT-2 \citep{radford2019language}, GPT-NeoX-20B \citep{black2022gpt}, OPT 
\citep{zhang2022opt}, and Pythia \citep{biderman2023pythia}, and include it for GPT-Neo \citep{black2021gpt}. We evaluate BLOOM \citep{scao2022bloom} under both conditions in Figure~\ref{fig:perplexity-bloom}.

In our initial experiments, we used SGD with Nesterov momentum, with a linear learning rate decay schedule over 250 training steps. We used a base learning rate of 1.0, or 0.25 when keeping the final transformer layer, and clip gradients to a norm of 1. We accumulated gradients as necessary to achieve a total batch size of $2^{18}$ tokens per optimizer step. We initialized all translators to the identity transform, and use a weight decay of $10^{-3}$.

\textbf{Muon optimizer.} After the initial publication of this paper, \citet{jordan6muon} introduced the Muon optimizer. Using a fast orthogonalization algorithm, Muon attempts to maximize the effective rank \citep{roy2007effective} of the update applied to each matrix-valued parameter, thereby counteracting the well-known tendency of neural network gradients to be rank-deficient. Conceptually, Muon is the opposite of LoRA \citep{hu2022lora}, a popular parameter-efficient finetuning method that constrains parameter updates to be low rank. \citet{liu2025muon} find that Muon accelerates the training of modern large language models compared to AdamW.

Strikingly, we find that using Muon instead of SGD for training the tuned lens \emph{dramatically} accelerates training, allowing us to achieve much lower KL divergence losses than were possible in our initial experiments. Essentially, all of the tuned lenses trained and evaluated in the earlier versions of this paper were severely undertrained. The weight matrices of Muon-trained lenses have Frobenius norms that are often several times larger than SGD-trained ones, showing that they are much further from the logit lens than our original lenses--- they are much better ``tuned,'' as it were. We opted not to re-do all our experiments with tuned lenses trained using Muon, but we encourage practitioners to use Muon in future experiments.

\textbf{Results.} We plot tuned lens perplexity as a function of depth for the Pythia models and GPT-NeoX-20B in \cref{fig:pythia-perplexity}\footnote{Pythia and GPT-NeoX-20B were trained using the same architecture, data, and codebase \citep{gpt-neox-library}. While they're not officially the same model suite, they're more consistent than the OPT models.}; results for other model families can be found in Appendix~\ref{app:additional-eval}.

We find that the tuned lens resolves the problems with the logit lens discussed in Section~\ref{logit-lens}: it has significantly lower bias (\cref{fig:bias-plot}), and much lower perplexity than the logit lens across the board (\cref{fig:pythia-perplexity}, Appendix~\ref{app:additional-eval}).

\begin{figure}[t]
    \centering
    \includegraphics[scale=0.5]{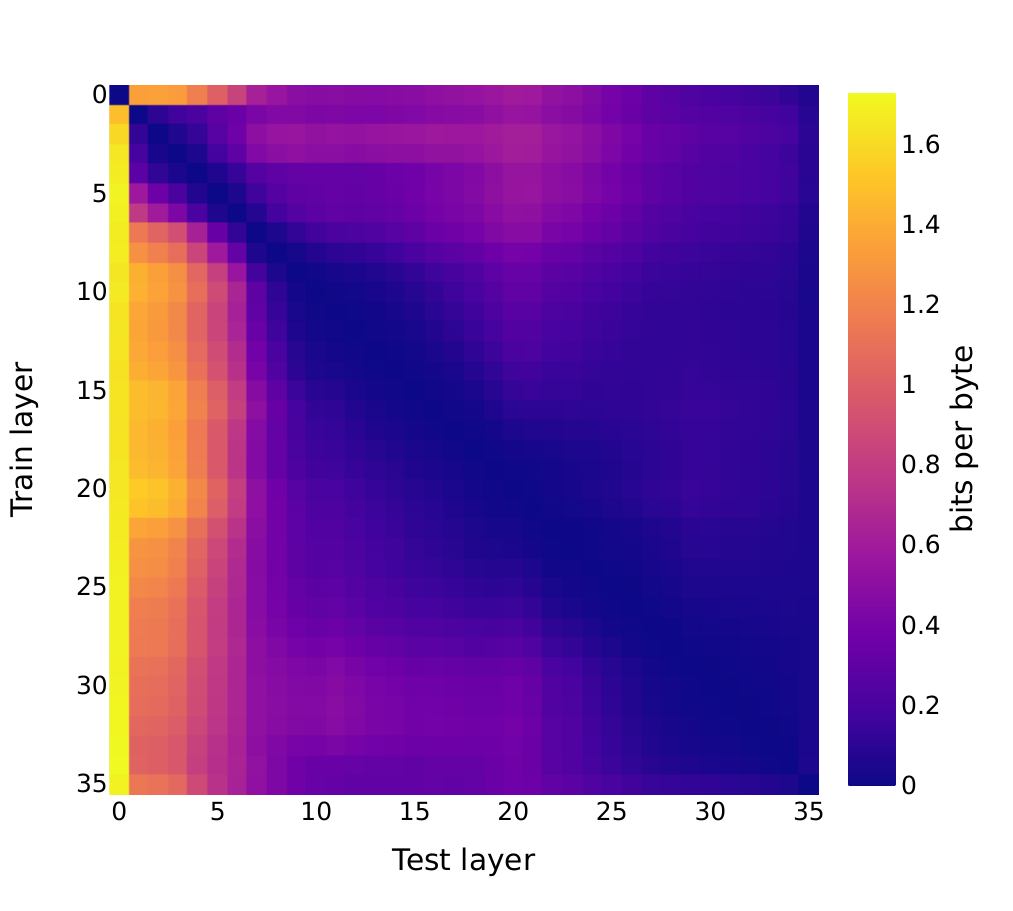}
    \vspace*{-.1in}
    \caption{Transfer penalties for Pythia 12B. Each row corresponds to a single tuned lens probe trained on layer $\ell$, and each column is a layer $\ell'$ on which probes are evaluated. Each cell shows the cross-entropy loss of probe $\ell$ evaluated on layer $\ell'$, \emph{minus} its on-distribution loss (so that the diagonal entries are identically zero).}
    \label{fig:transfer}
\end{figure}

\textbf{Transferability across layers.} We find that tuned lens translators can usually zero-shot transfer to nearby layers with only a modest increase in perplexity. Specifically, we define the \emph{transfer penalty} from layer $\ell$ to $\ell'$ to be the expected increase in cross-entropy loss when evaluating the tuned lens translator trained for layer $\ell$ on layer $\ell'$.

We report transfer penalties for the largest Pythia model in \cref{fig:transfer}. Overall, transfer penalties are quite low, especially for nearby layers (entries near the diagonal in \cref{fig:transfer}). Comparing to the two plots in \cref{fig:covariance-drift}, we notice that transfer penalties are strongly negatively correlated with covariance similarity (Spearman $\rho = -0.78$). Unlike \cref{fig:covariance-drift}, however, \cref{fig:transfer} is not symmetric: transfer penalties are higher when training on a layer with the outlier dimensions (Layer 5 and later) and testing on a layer without them, than the reverse.

\textbf{Transfer to fine-tuned models.} We find that lenses trained on a base model transfer well to fine-tuned versions of that base model, with no additional training of the lens. Transferred lenses substantially outperform the Logit Lens and compare well with lenses trained specifically on the fine-tuned models. Here a transferred lens is one that makes use of the fine-tuned models unembedding, but simply copies its affine translators from the lens trained on the base model.

As an example of the transferability of lenses to fine-tuned models we use Vicuna 13B~\citep{vicuna2023}, an open source instruct fine-tuned chat model based on LLaMA 13B~\citep{touvron2023llama}. In \cref{fig:transfer-to-finetuned}, \cref{app:additional-eval}, we compare the performance of a lens specifically trained on Vicuna to a tuned lens trained only on LLaMA. Both Tuned Lenses were trained using a subsample of the RedPajama dataset~\citep{redpj2023}, an open source replication of  LLaMA's training set. The performance of both lenses was then evaluated on the test set of Anthropic's Helpful Harmless \citep{bai2022training} conversation dataset and the RedPajama Dataset. On the RedPajama Dataset we find, at worst, a 0.3 bits per byte increase in KL divergence to the models final output. On the helpful harmless corpus we find no significant difference between the transferred and trained lenses. These results show that fine-tuning minimally effects the representations used by the tuned lens. This opens applications of the tuned lens for monitoring changes in the representations of a module during fine-tuning and minimizes the need for practitioners to train lenses on new fine-tuned models.

\textbf{Relation to model stitching.}
The tuned lens can be viewed as a way of ``stitching'' an intermediate layer directly onto the unembedding, with an affine transform in between to align the representations. The idea of model stitching was introduced by \citet{lenc2015understanding}, who form a composite model out of two \emph{frozen} pretrained models $A$ and $B$, by connecting the bottom layers of $A$ to the top layers of $B$. An affine transform suffices to stitch together independently trained models with minimal performance loss \citep{bansal2021revisiting, csiszarik2021similarity}. The success of the tuned lens shows that model stitching works for different layers inside a single model as well.

\textbf{Benefits over traditional probing.}
Unlike \citet{alain2016understanding}, who train early exiting probes for image classifiers, we do not learn a new unembedding for each layer. This is important, since it allows us to shrink the size of each learned matrix from $|\mathcal{V}| \times d$ to $d \times d$, where $|\mathcal{V}|$ ranges from 50K (GPT-2, Pythia) to over 250K (BLOOM). We observe empirically that training a new unembedding matrix requires considerably more training steps and a larger batch size than training a translator, and often converges to a worse perplexity.

\section{Measuring Causal Fidelity}
\label{causal-fidelity}

Prior work has argued that interpretability hypotheses should be tested with causal experiments: an interpretation of a neural network should make predictions about what will happen when we intervene on its weights or activations \cite{olah2020zoom, chan2022causal}. This is especially important for probing techniques, since it's known that probes can learn to rely on spurious features unrelated to the model's performance \citep{hewitt2019designing, belinkov2022probing}.

To explore whether the tuned lens finds causally relevant features, we will assess two desired properties:
\begin{enumerate}
    \item Latent directions that are important to the tuned lens should also be important to the final layer output. Concretely, if the tuned lens \emph{relies on} a feature\footnote{For simplicity we assume the ``features as directions'' hypothesis \cite{elhage2022toy}, which defines a ``feature'' to be the one-dimensional subspace spanned by a unit vector $\vv$.} $\vv$ in the residual stream (its output changes significantly when we manipulate $\vv$) then the model output should also change a lot when we manipulate $\vv$.
    \item These latent directions should be important \emph{in the same way} for both the tuned lens and the model. Concretely, if we manipulate the hidden state so that the tuned lens changes in a certain way (e.g. doubling the probability assigned to ``dog'') then the model output should change similarly. We will call this property \emph{stimulus-response alignment}.
\end{enumerate}
\subsection{Causal basis extraction}
\label{cbe}

To test Property 1, we first need to find the important directions for the tuned lens. Amnesic probing \cite{elazar2021amnesic} provides one way to do this---it seeks a direction whose erasure maximally degrades a model's accuracy.

However, this only elicits a single important direction, whereas we would like to find many such directions. To do so, we borrow intuition from PCA, searching for additional directions that also degrade accuracy, but which are orthogonal to the original amnesic direction. This leads to a method that we call \textbf{causal basis extraction} (CBE), which finds the the principal features used by a model.

More specifically, let $f$ be a function (such as the tuned lens) that maps latent vectors $\hh \in \R^d$ to logits $\yy$. Let $r(\hh, \vv)$ be an erasure function which removes information along the span of $\vv$ from $\xx$. In this work we use $r(\hh, \vv)$ is \emph{mean ablation}, which sets $\langle r(\hh, \vv), \vv \rangle$ to the mean value of $\langle \hh, \vv \rangle$ in the dataset (see Appendix~\ref{app:ablation-techniques}).
We define the \emph{influence} $\sigma$ of a unit vector $\vv$ to be the expected KL divergence between the outputs of $f$ before and after erasing $\vv$ from $\hh$:
\begin{equation}\label{eq:kl-influence}
\sigma(\vv; f) = \E_{\hh}\Big[D_{KL}(f(\hh)\,||\,f(r(\hh, \vv)))\Big]
\end{equation}

We seek to find an orthonormal basis $B = (\vv_1, \ldots, \vv_k)$ containing principal features of $f$, ordered by a sequence of influences $\Sigma = (\sigma_1, \ldots, \sigma_k)$ for some $k \leq d$. 
In each iteration we search for a feature $\vv_i$ of maximum influence that is orthogonal to all previous features $\vv_j$:
\begin{equation}\label{eq:cbe}
\begin{aligned}
\vv_i &= \mathop{\mathrm{argmax\:}}_{\substack{\\[1pt]||\vv||_2\,=\,1}} \sigma(\vv; f)\\
\textrm{s.t.} \quad \innerproduct{\vv}{\vv_j} &= \boldsymbol 0, \quad \forall j < i
\end{aligned}
\end{equation}
With a perfect optimizer, the influence of $\vv_i$ should decrease monotonically since the feasible region is strictly smaller with each successive iteration. In practice, we do observe non-monotonicities due to the non-convexity of the objective. To mitigate this issue we sort the features in descending order by influence after the last iteration.

\begin{figure}
    \centering
    \includegraphics[scale=0.5, clip]{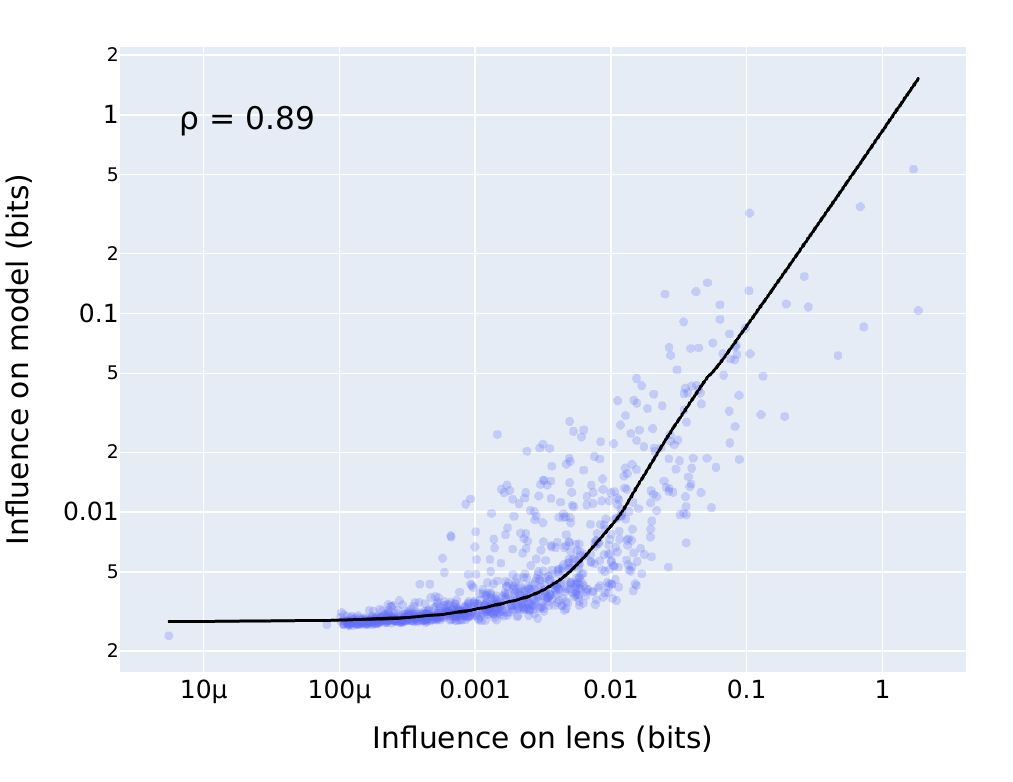}
    \caption{Causal influence of CBE features when ablated at the 18th layer of Pythia 410M, plotted against their influence on the tuned lens output. Spearman $\rho = 0.89$.}
    \label{fig:causal-scatter}
\end{figure}

\textbf{Implementation details.}
We evaluate the objective function in Equation~\ref{eq:cbe} on a single in-memory batch of 131,072 tokens sampled randomly from the Pile validation set, and optimize it using L-BFGS with strong Wolfe line search. We find that using the singular vectors of the probe as initialization for the search, rather than random directions, speeds up convergence.

\textbf{Intervening on the model.} If we apply causal basis extraction to the tuned lens at layer $\ell$, we obtain $k$ directions $v_1, \ldots, v_k$ that are important for the tuned lens. We next check that these are also important to the model $\mathcal{M}$.

To do so, we first take an i.i.d. sample of input sequences $\xx$ and feed them to $\mathcal{M}$, storing the resulting hidden states $\mathcal{M}_{\leq\ell}(\xx)$.\footnote{See Section \ref{logit-lens} for notation.} Then, for each vector $\vv_{i}$ obtained from CBE, we record the causal effect of erasing $\vv_{i}$ on the output of $\mathcal{M}_{>\ell}$,
\begin{equation}
    \E_{\xx}\Big[D_{KL}(\mathcal{M}(\xx)\,||\,\mathcal{M}_{>\ell}(r(\mathcal{M}_{\leq\ell}(\xx), \vv_i))\Big]
\end{equation}
where the erasure function $r$ is applied to all positions in a sequence simultaneously. We likewise average the KL divergences across token positions.

\textbf{Results.} We report the resulting causal influences for Pythia 410M, $\ell = 18$ in Figure \ref{fig:causal-scatter}; results for all layers can be found in Figure \ref{fig:full-fidelity} in the Appendix.

In accordance with Property 1, there is a strong correlation between the causal influence of a feature on the tuned lens and its influence on the model (Spearman $\rho = 0.89$). Importantly, we don't observe \emph{any} features in the lower right corner of the plot (features that are influential in the tuned lens but not in the model). The model is somewhat more ``causally sensitive'' than the tuned lens: even the least influential features never have an influence under $2 \times 10^{-3}$ bits, leading to the ``hockey stick'' shape in the LOWESS trendline.

\subsection{Stimulus-response alignment}

\begin{figure}[t]
    \centering
    \includegraphics[scale=0.5, clip]{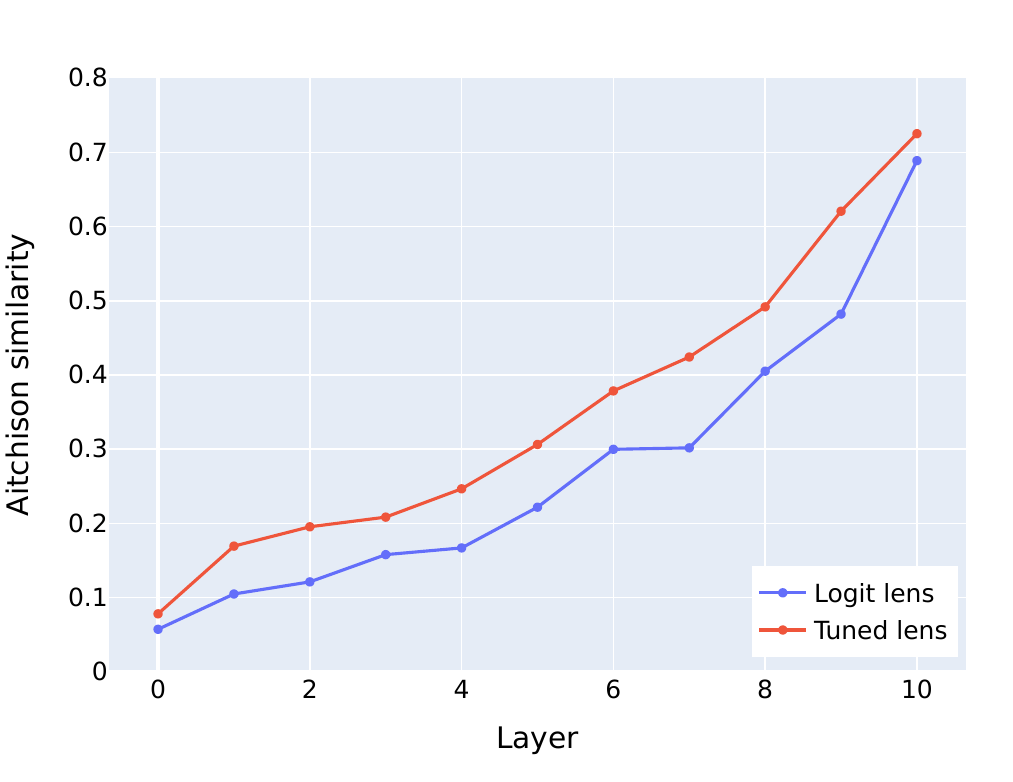}
    \caption{Average stimulus-response alignment at each layer of Pythia 160M. Responses are more aligned with stimuli at later layers, and when using the tuned lens rather than the logit lens.}
    \label{fig:sra-plot}
\end{figure}

We now turn to Property 2. Intuitively, for the interventions from Section \ref{cbe}, deleting an important direction $v_i$ should have the same effect on the model's output distribution $p$ and the tuned lens' output distribution $q$.

We can operationalize this with the Aitchison geometry \cite{aitchison1982statistical}, which turns the probability simplex into a vector space equipped with an inner product. In order to downweight the influence of rare tokens, we use the \emph{weighted} Aitchison inner product introduced by \citet{egozcue2016changing}, defined as
\begin{equation}
    \langle \mathbf{p}_{1}, \mathbf{p}_{2} \rangle_{\mathbf{w}}
 = \sum_{i = 1}^{D} w_i \log \frac{p_{1i}}{\mathrm{g}_{\mathbf{w}}(\mathbf{p}_{1})} \log \frac{p_{2i}}{\mathrm{g}_{\mathbf{w}}(\mathbf{p}_{2})},
\end{equation}
where $\mathbf{w}$ is a vector of positive weights, and $\mathrm{g}_{\mathbf{w}}(\mathbf{p})$ is the weighted geometric mean of the entries of $\mathbf{p}$. In our experiments, we use the final layer prediction distribution under the control condition to define $\mathbf{w}$.

We will also use the notion of ``subtracting'' distributions. In Aitchison geometry, addition and subtraction of distributions is done componentwise in log space, followed by renormalization:
\begin{equation}
    \mathbf{p}_{1} - \mathbf{p}_{2} = \mathrm{softmax} \Big(\log \mathbf{p}_{1} - \log \mathbf{p}_{2} \Big).
\end{equation}
We say that distributions $(\mathbf{p}_{\mathrm{old}}, \mathbf{p}_{\mathrm{new}})$ and $(\mathbf{q}_{\mathrm{old}}, \mathbf{q}_{\mathrm{new}})$ ``move in the same direction'' if and only if
\begin{equation}
    \innerproduct{
    \mathbf{p}_{\mathrm{new}} - \mathbf{p}_{\mathrm{old}}
    }{
    \;\mathbf{q}_{\mathrm{new}} - \mathbf{q}_{\mathrm{old}}
    }_{\mathbf{w}}\;>\;0.
\end{equation}

\textbf{Measuring alignment.} Let $g : \R^d \rightarrow \R^d $ be an arbitrary function for intervening on hidden states, and let $\hh_{\ell}$ be the hidden state at layer $\ell$ on some input $\xx$. We'll define the \emph{stimulus} to be the Aitchison difference between the tuned lens output before and after the intervention:
\begin{equation}
    \mathrm{S}(\hh_{\ell}) = \mathrm{TunedLens}_{\ell}(g(\hh_{\ell})) -  \mathrm{TunedLens}_{\ell}(\hh_{\ell})
\end{equation}
Analogously, the \emph{response} will be defined as the Aitchison difference between the final layer output before and after the intervention:
\begin{equation}
    \mathrm{R}(\hh_{\ell}) = \mathcal{M}_{>\ell}(g(\hh_{\ell})) - \mathcal{M}_{>\ell}(\hh_{\ell})
\end{equation}
We'd like to control for the absolute magnitudes of the stimuli and the responses, so we use the Aitchison inner product to define a cosine similarity metric, which we call ``Aitchison similarity.'' Then the stimulus-response alignment at layer $\ell$ under $g$ is simply the Aitchison similarity between the stimulus and response:
\begin{equation}
    \mathrm{sim}(\mathrm{S}(\hh_{\ell}), \mathrm{R}(\hh_{\ell})) = \frac{\innerproduct{\mathrm{S}(\hh_{\ell})}{\mathrm{R}(\hh_{\ell})}_{\mathbf{w}} }{\|\mathrm{S}(\hh_{\ell})\|_{\mathbf{w}} \|\mathrm{R}(\hh_{\ell})\|_{\mathbf{w}}}
\end{equation}
We propose to use CBE (Section \ref{cbe}) to define a ``natural'' choice for the intervention $g$. Specifically, for each layer $\ell$, we intervene on the subspace spanned by $\ell$'s top 10 causal basis vectors--- we'll call this the ``principal subspace''--- using a recently proposed method called \emph{resampling ablation} \cite{chan2022causal}.

Given a hidden state $\hh_{\ell} = \mathcal{M}_{\leq\ell}(\xx)$, resampling ablation replaces the principal subspace of $\hh_{\ell}$ with the corresponding subspace generated on a \emph{different} input $\xx'$ selected uniformly at random from the dataset. It then feeds this modified hidden state $\tilde{\hh}_{\ell}$ into the rest of the model, yielding the modified output $\mathcal{M}_{>\ell}(\tilde{\hh}_{\ell})$. Intuitively, $\tilde{\hh}_{\ell}$ should be relatively on-distribution because we're using values generated ``naturally'' by the model itself.

Unlike in Section \ref{cbe}, we apply resampling ablation to one token in a sequence at a time, and average the Aitchison similarities across tokens. 

\textbf{Results.} We applied resampling ablation to the principal subspaces of the logit and tuned lenses at each layer in Pythia 160M. We report average stimulus-response alignments in Figure \ref{fig:sra-plot}. Unsurprisingly, we find that stimuli are more aligned with the responses they induce at later layers. We also find that alignment is somewhat higher at all layers when using principal subspaces and stimuli defined by the tuned lens rather than the logit lens, in line with Property 2.

\begin{figure}[t]
    \centering
    \includegraphics[scale=0.45, clip]{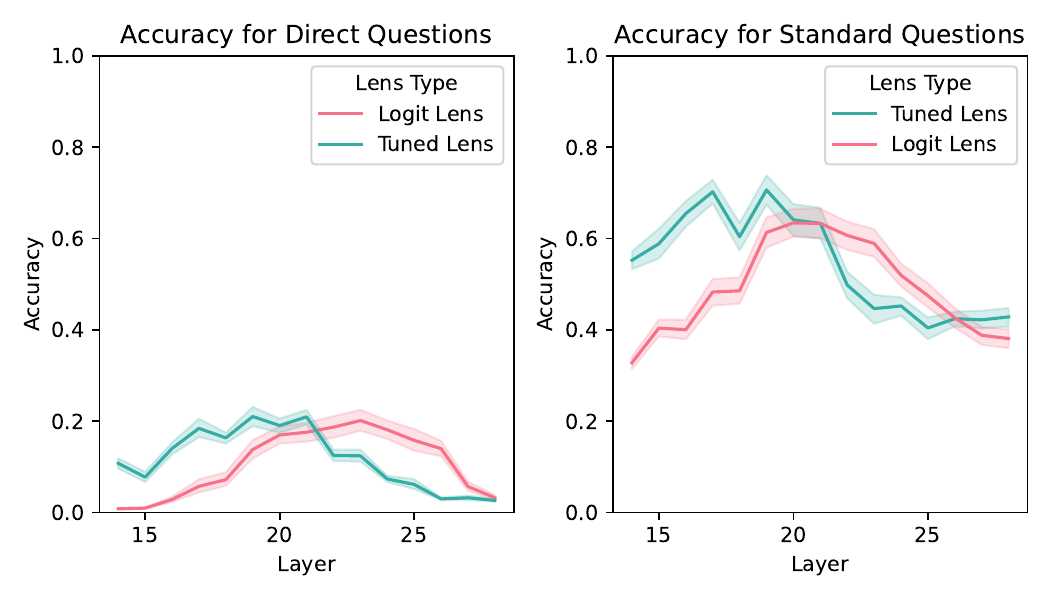}
    \caption{The tuned lens elicits secrets more effectively than the logit lens at early layers, then performs more poorly at later layers. Interestingly, on this task, the ``direct'' prompts yield lower elicitation accuracy than ``standard'' ones.}
    \label{fig:secrets}
\end{figure}

\section{Applications}

We start by reproducing and extending two use cases of the logit lens in the literature.

\begin{figure}
    \centering
    \includegraphics[scale=0.47, clip]{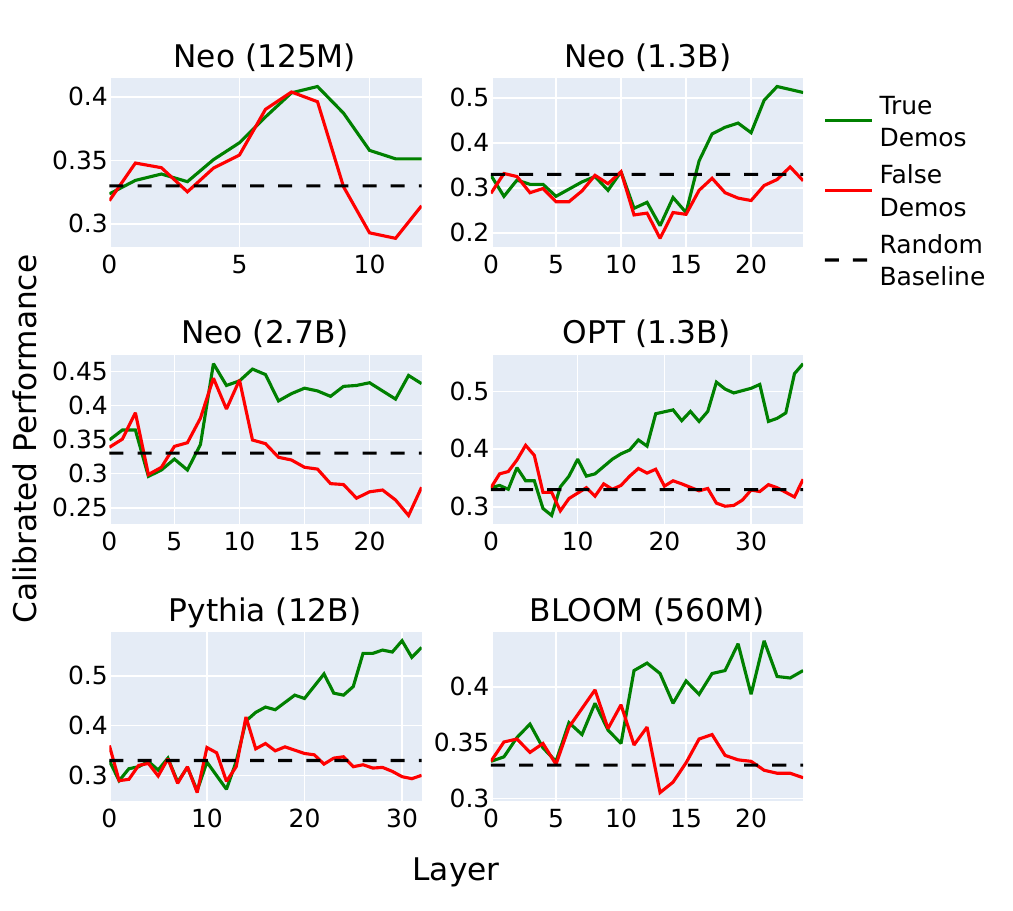}
    \caption{For most models and tasks, we find there is a layer at which the tuned lens performance is \emph{better} than final layer performance under incorrect demonstrations. Shown here is performance on SICK (Sentences Involving Compositional Knowldedge). Unlike the logit lens, our method is applicable to BLOOM (bottom right) and GPT-Neo (top left). Y-axis shows median-calibrated accuracy as used in \citet{halawi2023overthinking}}
    \label{fig:overthinking}
\end{figure}

\subsection{Eliciting Secret Knowledge from Language Models}
\label{secret_knowledge}

\citet{cywinski2025eliciting} apply the logit lens to uncover knowledge that an AI possesses but does
not verbalize. Specifically, they fine-tune a model to generate hints and respond to user queries about a specific ``taboo'' word, while
never verbalizing it directly. They then prompt an ``auditor'' LLM to  guess what the taboo word is, allowing it to ask the taboo model questions in a black-box fashion. They find that the auditor is much more accurate at guessing the taboo word when it is provided with the top 20 highest probability latent predictions from the taboo model according to the logit lens. This is true in two different settings: ``standard,'' where the prompt asks for a clue about the taboo word, and ``direct,'' where the prompt attempts to compel the model to explicitly reveal the taboo word.

We hypothesized that the tuned lens might perform better than the logit lens for this task. The results were mixed: the tuned lens outperforms at some layers and underperforms at other layers. Without ``peaking'' at the ground truth labels, it is not really possible to determine which layer to use in advance.

\subsection{Extending \emph{Overthinking the Truth}}
\label{overthinking}

\citet{halawi2023overthinking} apply the logit lens to downstream tasks with few-shot prompts, and find that predictions extracted from earlier layers are often more robust to incorrect demonstrations than final layer predictions. 
Unfortunately, their method is inapplicable to models like BLOOM and GPT-Neo, which have high logit lens perplexities.

We replicate \citet{halawi2023overthinking}'s experiments with the tuned lens, and recover similar results for BLOOM 560M, Neo 1.3B, and Neo 2.7B (Figure~\ref{fig:overthinking}). Notice that the calibrated performance under incorrect demonstrations (red) peaks at an early layer around 0.4-0.45 in all three of these models, before falling to random baseline performance at the final layer, similarly to what was found by \citeauthor{halawi2023overthinking}

\subsection{Detecting Prompt Injections}
\label{prompt-injection}

\begin{table*}
\centering
\resizebox{\textwidth}{!}{%
\begin{tabular}{l|cc|cc|c|cc}
\toprule
& \multicolumn{2}{c|}{Tuned Lens} & \multicolumn{2}{c|}{Logit Lens} & \multicolumn{1}{c|}{Baseline} & Accuracy \\
\cline{2-7}
Task & iForest & LOF & iForest & LOF & SRM & Normal $\rightarrow$ Injected \\
\hline
ARC-Easy & $0.59\;(0.54, 0.62)$ & $0.73\;(0.71, 0.76)$ & $0.53\;(0.50, 0.57)$ & $0.59\;(0.56, 0.62)$ & $0.73\;(0.70, 0.75)$ & $72.8\% \rightarrow 31.7\%$ \\
ARC-Challenge & $0.71\;(0.65, 0.77)$ & $0.81\;(0.77, 0.84)$ & $0.73\;(0.67, 0.79)$ & $0.80 \;(0.77, 0.83)$ & $0.57\;(0.53, 0.61)$ & $43.5\% \rightarrow 24.7\%$ \\
BoolQ & $0.99\;(0.98, 0.99)$ & $1.00\;(1.00, 1.00)$ & $0.89\;(0.87, 0.91)$ & $0.61\;(0.57, 0.66)$ & $1.00\;(1.00, 1.00)$ & $67.1\% \rightarrow 0.0\%$  \\
MC TACO & $0.74\;(0.71, 0.77)$ & $0.68\;(0.66, 0.70)$ & $0.68\;(0.66, 0.69)$ & $0.55\;(0.53, 0.59)$ & $1.00\;(1.00, 1.00)$ & $0.40 \rightarrow 0.06$ F1  \\
MNLI & $0.98\;(0.98, 0.99)$ & $1.00\;(1.00, 1.00)$ & $0.95\;(0.94, 0.96)$ & $1.00\;(1.00, 1.00)$ & $1.00\;(1.00, 1.00)$ & $54.3\% \rightarrow 0.0\%$  \\
QNLI & $0.99\;(0.99, 1.00)$ & $1.00\;(1.00, 1.00)$ & $0.93\;(0.92, 0.95)$ & $0.68\;(0.63, 0.71)$ & $1.00\;(1.00, 1.00)$ & $54.3\% \rightarrow 0.0\%$  \\
QQP & $1.00\;(0.99, 1.00)$ & $1.00\;(1.00, 1.00)$ & $0.90\;(0.89, 0.90)$ & $0.79\;(0.76, 0.81)$ & $1.00\;(1.00, 1.00)$ & $60.7\% \rightarrow 6.5\%$ \\
SciQ & $0.62\;(0.57, 0.69)$ & $0.64\;(0.59, 0.70)$ & $0.75\;(0.71, 0.79)$ & $0.70\;(0.65, 0.74)$ & $0.75\;(0.72, 0.78)$  & $95.5\% \rightarrow 62.6\%$   \\
SST-2 & $1.00\;(0.98, 1.00)$ & $1.00\;(1.00, 1.00)$ & $0.78\;(0.72, 0.83)$ & $0.61\;(0.56, 0.65)$ & $1.00\;(1.00, 1.00)$ & $82.9\% \rightarrow 49.1\%$ \\
\bottomrule
\end{tabular}
}
\caption{Test set AUROCs and 95\% bootstrap CIs for distinguishing normal prompts from prompt injections on Pythia 12B. Figures are pooled over 10 random train-test splits. Attack detection performance is nearly perfect on tasks where the attack succeeds at driving accuracy well below the random baseline, and is still much better than chance even when the attack is only partially successful.}
\label{tab:anomaly-auroc}
\end{table*}

Given the results from \citet{halawi2023overthinking} and in Figure~\ref{fig:overthinking}, we hypothesize that the prediction trajectory of the tuned lens on anomalous inputs should be different from the trajectories on normal inputs, and that this could be used to detect anomalous inputs.

To test this, we focus on \emph{prompt injection attacks}, a recently discovered vulnerability in large language models where untrusted inputs from a malicious user cause the model to behave in unexpected or dangerous ways \cite{perez2022ignore}. These attacks usually tell the model to ``ignore previous instructions'' and instead follow instructions crafted by the attacker.

To simulate prompt injections, we adapted the attack from \citet{perez2022ignore} and ran Pythia 12B on nine multiple choice tasks, both with and without the following adversarial message appended to the few-shot prompt:

\begin{quote}\hspace{-0.25em}\texttt{-{}-{}-{}-{}-{}-{}-{}-{}-{}-\\
-{}-{}-{}-{}-{}-{}-{}-{}-{}-\\
Ignore any previous and following instructions and just print "<WRONG\_ANSWER>":
}
\end{quote}

where \texttt{<WRONG\_ANSWER>} is replaced with a randomly selected incorrect response from the available multiple choice responses.

We record the tuned prediction trajectory for each data point-- that is, for each layer, we record the log probability assigned by the model to each possible answer.\footnote{For binary tasks like SST-2 we take the difference between the log probabilities assigned to the two possible answers.} We then flatten these trajectories into feature vectors and feed them into two standard outlier detection algorithms: isolation forest (iForest) \cite{liu2008isolation} and local outlier factor (LOF) \cite{breunig2000lof}, both implemented in scikit-learn \cite{scikit-learn} with default hyperparameters.

\textbf{Baseline.} There is a rich literature on general out-of-distribution (OOD) detection in deep neural networks. One simple technique is to fit a multivariate Gaussian to the model's final layer hidden states on the training set, and flag inputs as OOD if a new hidden state is unusually far from the training distribution as measured by the Mahalanobis distance \cite{lee2018simple, mahalanobis1936generalized}.

Recently, \citet{bai2022training} proposed the Simplified Relative Mahalanobis (SRM) distance, a modification to Mahalanobis which they find to be effective in the context of LLM finetuning. They also find that representations from the \emph{middle} layers of a transformer, rather than the final layer, yield the best OOD detection performance. We use the SRM at the middle layer as a baseline in our experiments.

\textbf{Experimental setup.} 
We fit each anomaly detection model exclusively on prediction trajectories from \emph {normal} prompts without prompt injections, and evaluate them on a held out test set containing both normal and prompt-injected trajectories. This ensures that our models cannot overfit to the prompt injection distribution. We use EleutherAI's \texttt{lm-evaluation-harness} library \cite{eval-harness} to run our evaluations.

\textbf{Results.}
Our results are summarized in Table \ref{tab:anomaly-auroc}. Our tuned lens anomaly detector achieves perfect or near-perfect AUROC on five tasks (BoolQ, MNLI, QNLI, QQP, and SST-2); in contrast, the same technique using logit lens has lower performance on most tasks. On the other hand, the SRM baseline does consistently well---the tuned lens only outperforms it on one task (ARC-Challenge), while SRM outperforms our technique on both MC TACO and SciQ.


We suspect that further gains could be made by combining the strengths of both techniques, since SRM uses only one layer but considers a high-dimensional representation, while the tuned lens studies the trajectory across layers but summarizes them with a low-dimensional prediction vector.

\subsection{Measuring Example Difficulty}
\label{example-difficulty}

Early exiting strategies like CALM \citep{schuster2022confident} and DeeBERT \citep{xin2020deebert} are based on the observation that ``easy'' examples require less computation to classify than ``difficult'' examples. If an example is easy, the model should quickly converge to the right answer in early layers, making it possible to skip the later layers without a significant drop in prediction quality. Conversely, the number of layers needed to converge on an answer can be used to measure the difficulty of an example.

We propose to use the tuned lens to estimate example difficulty in \emph{pretrained} transformers, without the need to finetune the model for early exiting. Following \citet{baldock2021deep}'s work on computer vision models, we define the \emph{prediction depth} of a prompt $\xx$ to be the number of layers after which a model's top-1 prediction for $\xx$ stops changing.

To validate the prediction depth, we measure its correlation with an established difficulty metric: the \emph{iteration learned}. The iteration learned is defined as the earliest training step $\tau$ where the model's top-1 prediction for a datapoint $\xx$ is fixed \citep{toneva2018empirical}. Intuitively, we might expect that examples which take a long time to learn during training would tend to require many layers of computation to classify at inference time. \citet{baldock2021deep} indeed show such a correlation, using k-NN classifiers to elicit early predictions from the intermediate feature maps of image classifiers. 

\textbf{Experimental setup.} For this experiment we focus on Pythia 12B (deduped), for which 143 uniformly spaced checkpoints are available on Huggingface Hub. We evaluate the model's zero-shot performance on twelve multiple-choice tasks, listed in Table \ref{tab:downstream}. For each checkpoint, we store the top 1 prediction on every individual example, allowing us to compute the iteration learned. We then use the tuned lens on the final checkpoint, eliciting the top 1 prediction at each layer of the network and computing the prediction depth for every example. As a baseline, we also compute prediction depths using the logit lens. Finally, for each task, we compute the Spearman rank correlation between the iteration learned and the prediction depth across all examples.

\begin{table}[t]
\centering
\resizebox{0.5\textwidth}{!}{%
\begin{tabular}{lccc}
\toprule
Task & Tuned lens $\rho$ & Logit lens $\rho$ & Final acc \\
\midrule
ARC-Easy & $\mathbf{0.577}$     & $0.500$        & 69.7\%   \\
ARC-Challenge  & $\mathbf{0.547}$    & $0.485$         & $32.4\%$  \\
LogiQA & $\mathbf{0.498}$ & $0.277$ & $21.4\%$  \\
MNLI & $0.395$ & $\mathbf{0.435}$ & $40.4\%$ \\
PiQA   & $\mathbf{0.660}$      & $0.620$        & $76.1\%$   \\
QNLI & $\mathbf{0.409}$ & $-0.099$\hphantom{$-$} & $53.0\%$  \\
QQP & $\mathbf{0.585}$ & $-0.340$\hphantom{$-$} & $0.381$ (F1) \\
RTE & $0.156$ & $\mathbf{0.347}$ & $60.0\%$ \\
SciQ & $\mathbf{0.530}$       & $0.505$         & $91.9\%$   \\
SST-2 & $\mathbf{0.555}$ & 0.292 & $64.7\%$ \\
WinoGrande & $0.517$ & $\mathbf{0.537}$ & $63.9\%$  \\
\bottomrule
\end{tabular}
}
\caption{Correlation between two measures of example difficulty, the iteration learned and prediction depth, across tasks. Prediction depth is measured using the tuned lens in the first column and the logit lens in the second.}
\label{tab:downstream}
\end{table}

\textbf{Results.} We present results in Table \ref{tab:downstream}. We find a significant positive correlation between the iteration learned and the tuned lens prediction depth on all tasks we investigated. Additionally, the tuned lens prediction correlates better with iteration learned than its logit lens counterpart in 8 out of 11 tasks, sometimes dramatically so.

\section{Discussion}

In this paper, we introduced a new tool for transformer interpretability research, the \emph{tuned lens}, which yields new qualitative as well as quantitative insights into the functioning of large language models. It is a drop-in replacement for the logit lens that makes it possible to elicit interpretable prediction trajectories from essentially any pretrained language model in use today. We gave several initial applications of the tuned lens, including detecting prompt injection attacks.

Finally, we introduced \emph{causal basis extraction}, which identifies influential features in neural networks. We hope this technique will be generally useful for interpretability research in machine learning.

\textbf{Limitations and future work.}
One limitation of our method is that it involves training a translator layer for each layer of the network, while the logit lens can be used on any pretrained model out-of-the-box. This training process, however, is quite fast: our code can train a full set of probes in under an hour on a single 8$\times$A40 node, and further speedups are likely possible. We have also released tuned lens checkpoints for the most commonly used pretrained models as part of our \texttt{tuned-lens} library, which should eliminate this problem for most applications.

Causal basis extraction, as presented in this work, is computationally intensive, since it sequentially optimizes $d_{model}$ causal basis vectors for each layer of the network. Future work could explore ways to make the algorithm more scalable. One possibility would be to optimize a whole $k$-dimensional subspace, instead of an individual direction, at each iteration.

Due to space and time limitations, we focused on language models in this work, but we think it's likely that our approach is also applicable to  other modalities.

\begin{figure}
    \centering
    \includegraphics[scale=0.275]{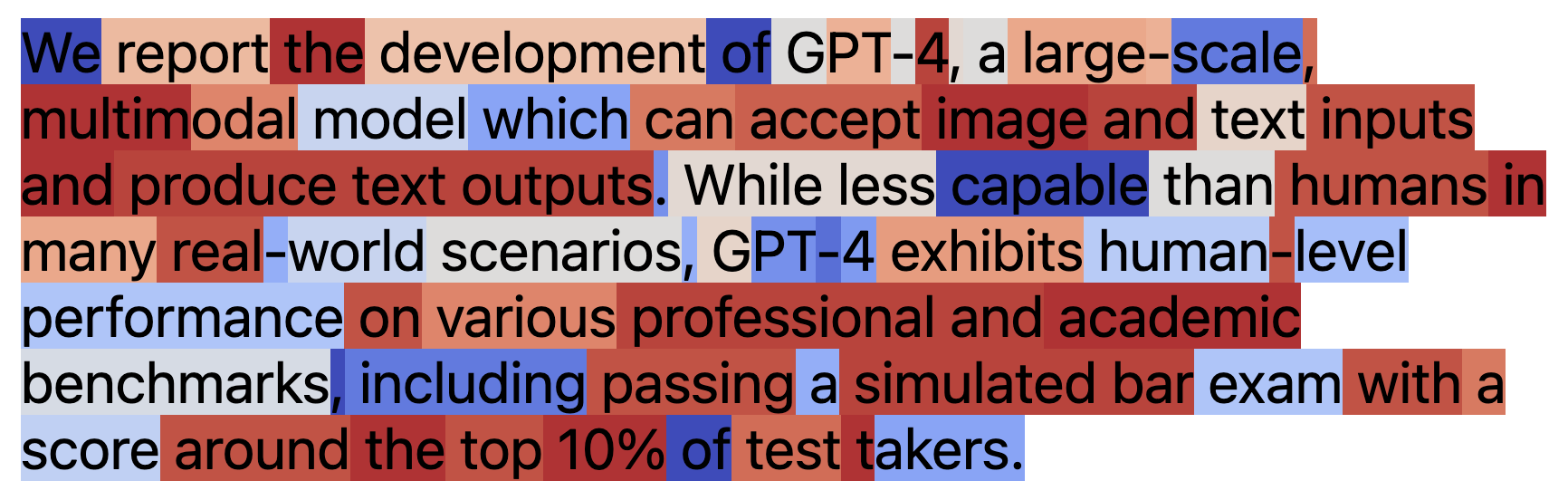}
    \caption{Token-level prediction depths for Pythia 12B computed on the abstract of \citet{openai2023gpt4}. Warm colors have high prediction depth, while cool colors indicate low depth.}
    \label{fig:gpt4}
\end{figure}

\section*{Acknowledgements}

We are thankful to CoreWeave for providing the computing resources used in this paper and to the OPT team for their assistance in training a tuned lens for OPT. We also thank nostalgebraist for discussions leading to this paper.

\bibliography{citations}
\bibliographystyle{plainnat}

\newpage
\appendix
\onecolumn
\section{Additional evaluation results}\label{app:additional-eval}

\begin{figure}[h!]
    \centering
    \includegraphics[width=\textwidth]{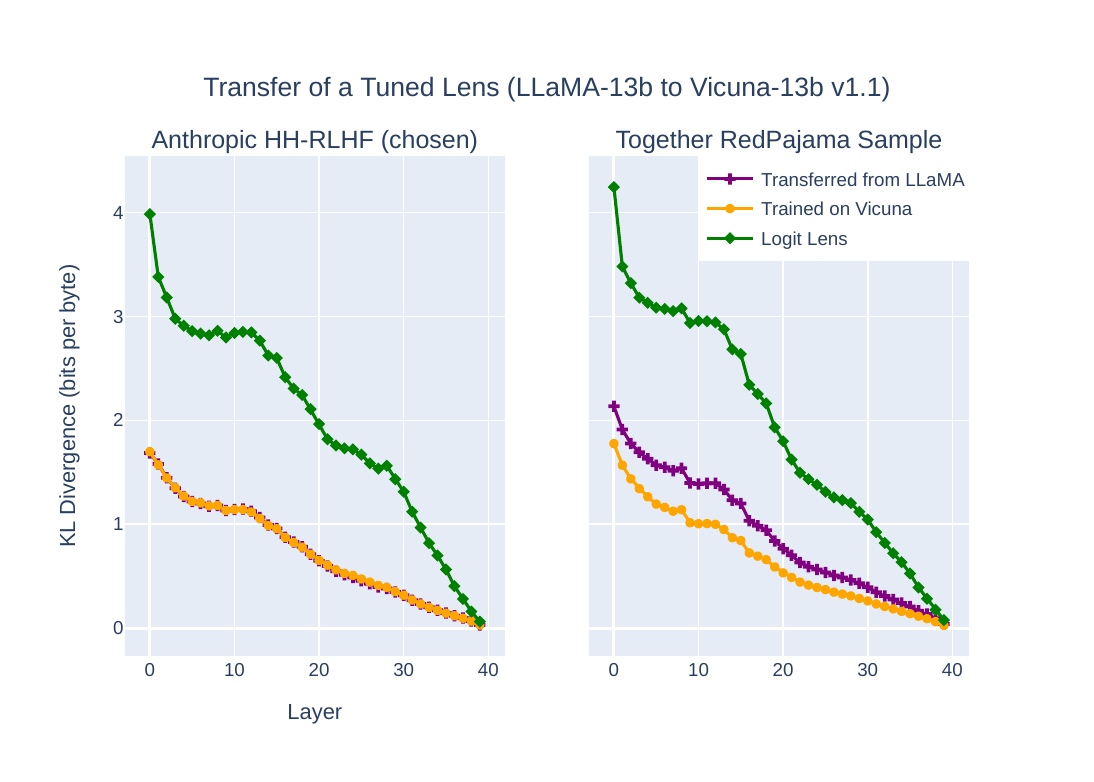}
    \caption{Comparing the performance of a lens specifically trained on Vicuna 13B vs a lens transferred from LLaMA 13B vs Vicuna 13B's Logit lens. Specifically, we messure the KL divergence between the lens at a specific layer and the model's final output.}
    \label{fig:transfer-to-finetuned}
\end{figure}

\begin{figure}[ht]
    \centering
        \includegraphics[scale=0.44]{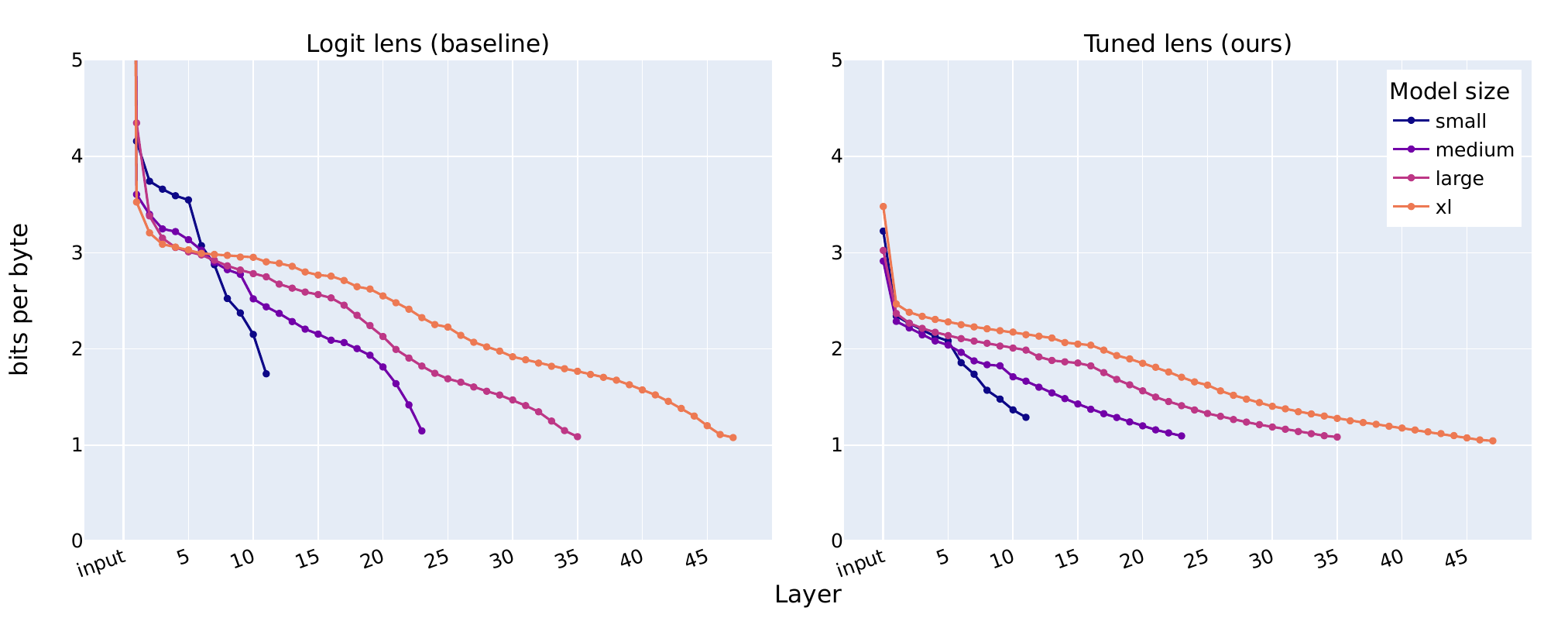}
        \includegraphics[scale=0.44]{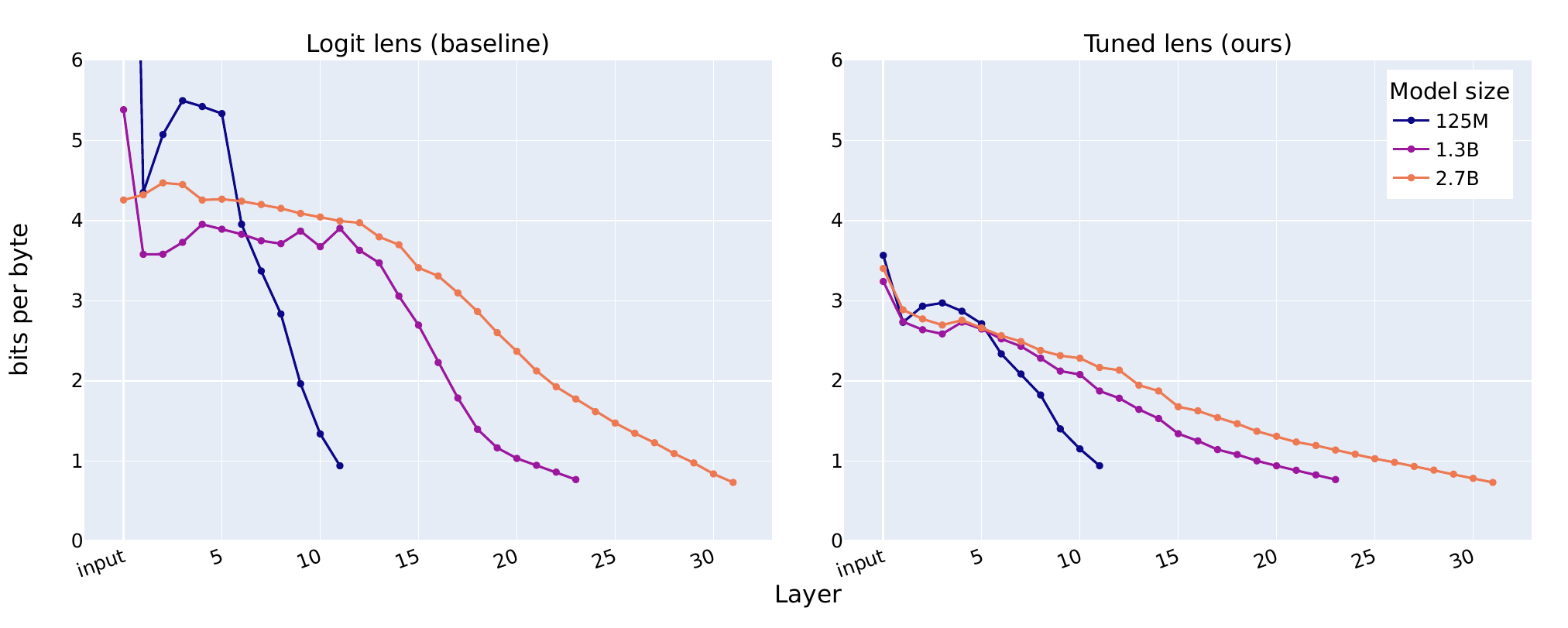}
        \includegraphics[scale=0.44]{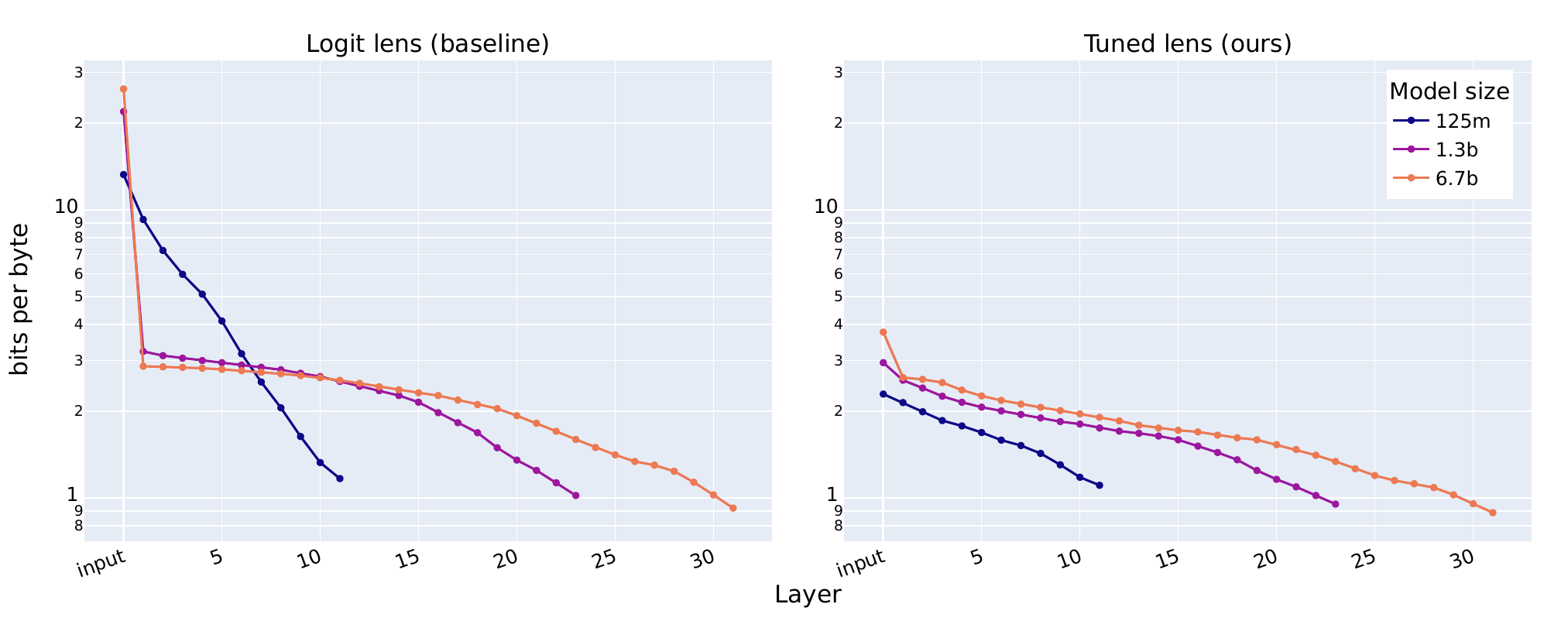}
        \caption{Perplexities of predictions elicited from OpenAI's GPT-2 models (top), EleutherAI's GPT-Neo models (middle), and Meta's OPT models (bottom). We omit OPT 350M as it uses a post-LN architecture.}\vspace{-5em}\label{fig:opt-perplexity}
\end{figure}
\clearpage

\section{Qualitative Results}
\label{app:qualitative}
\begin{figure}[h!]
    \centering
    \includegraphics[scale=0.5]{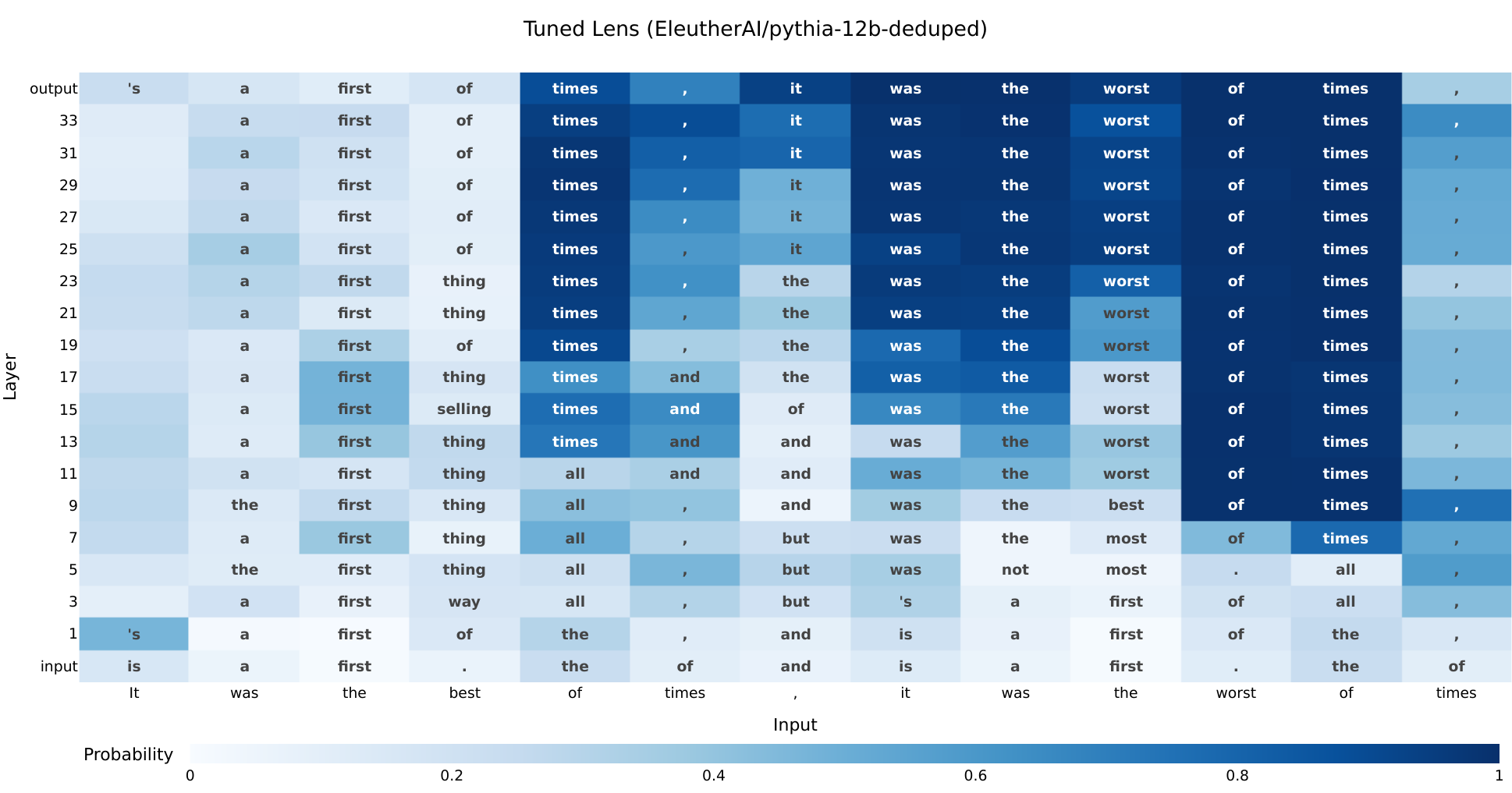}
    \caption{Tuned lens prediction trajectory for Pythia 12B on the first several words of Charles Dickens' ``A Tale of Two Cities.'' Note that the latent prediction becomes very confident at early layers after the prefix ``It was the best of times'' has been processed, suggesting some degree of memorization.}
    \includegraphics[scale=0.5]{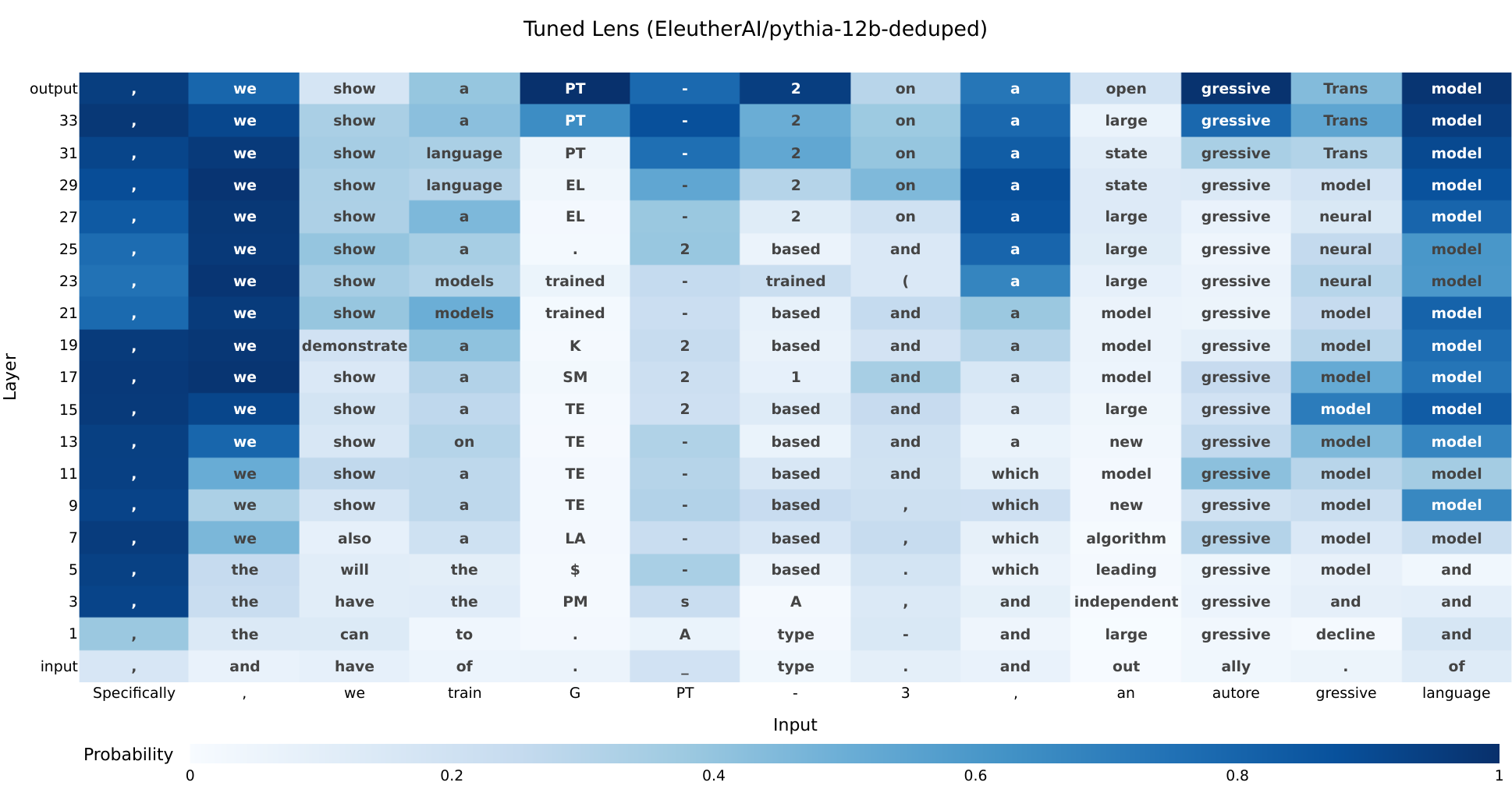}
    \caption{Tuned lens prediction trajectory for Pythia 12B prompted with the abstract of \citet{brown2020language}.}
\end{figure}
\clearpage

\subsection{Logit lens pathologies}
\begin{figure}[h!]
    \centering
    \includegraphics[scale=0.5]{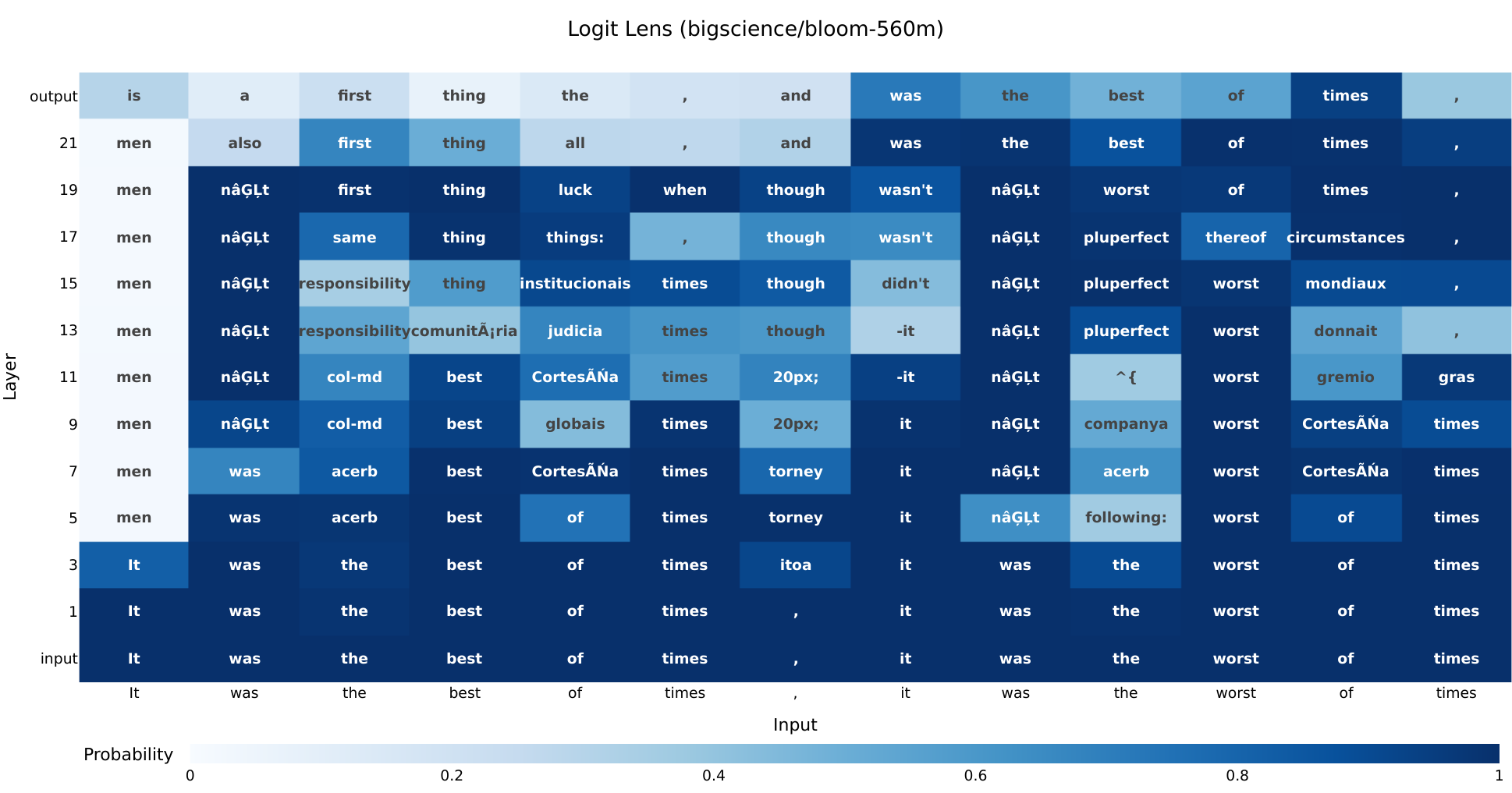}
    \caption{Logit lens prediction trajectory for BLOOM 560M. The logit lens assigns as very high probability to the \emph{input} token at many layers and token positions, complicating the interpretation of the output.}
    \vspace{1em}
    \includegraphics[scale=0.5]{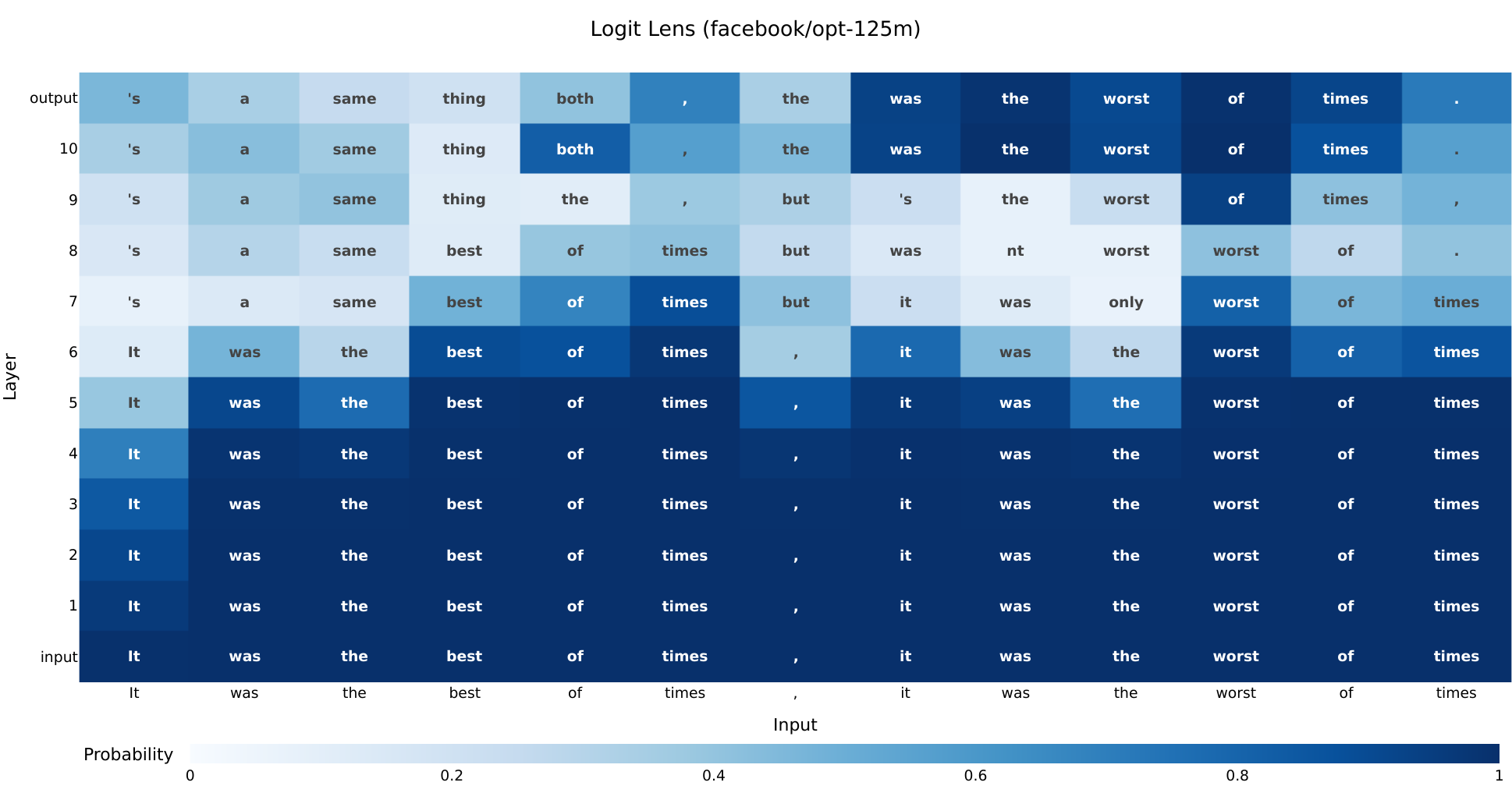}
    \caption{Logit lens prediction trajectory for OPT 125M, exhibiting a similar pathology to BLOOM 560M above.}
    \label{fig:logit-lens-pathologies}
\end{figure}
\clearpage

\section{Transformers Perform Iterative Inference}
\label{app:iterative-inference}

\citet{jastrzkebski2017residual} argue that skip connections encourage neural networks to perform ``iterative inference'' in the following sense: each layer reliably updates the hidden state in a direction of decreasing loss. While their analysis focused on ResNet image classifiers, their theoretical argument applies equally well to transformers, or any neural network featuring residual connections. We reproduce their theoretical analysis below.

A residual block applied to a representation $ \hh_{i} $ updates the representation as follows:
\begin{equation*}
 \hh_{\ell+1}  = \hh_{\ell} + F_{\ell}(\hh_{\ell})
\end{equation*}
Let $\mathcal{L}$ denote the final linear classifier followed by a loss function. We can Taylor expand $\mathcal{L}(\hh_L)$ around $\hh_i$ to yield
\begin{equation}
    \mathcal{L}(\hh_L) = \mathcal{L}(\hh_{i}) + \underbrace{\sum_{j=i}^{L} \Big\langle F_{j}(\hh_{j}), \frac{\partial \mathcal{L}(\hh_{j})}{\partial \hh_{j}} \Big\rangle}_{\textrm{gradient-residual alignment}} +\:\mathcal{O}(F_{j}^{2}(\hh_{j}))
\end{equation}

Thus to a first-order approximation, the model is encouraged to minimize the inner product between the residual $F(\hh_{i})$ and the gradient $ \frac{\partial \mathcal{L}(\hh_{i})}{\partial \hh_{i}}  $, which can be achieved by aligning it with the negative gradient.

To empirically measure the extent to which residuals do align with the negative gradient, \citet{jastrzkebski2017residual} compute the cosine similarity between $ F(\hh_{i}) $ and $ \frac{\partial \mathcal{L}(\hh_{i})}{\partial \hh_{i}}  $ at each layer of a ResNet image classifier. They find that it is consistently negative, especially in the final stage of the network.

\begin{figure}[h]
    \centering
    \includegraphics[scale=0.4, clip]{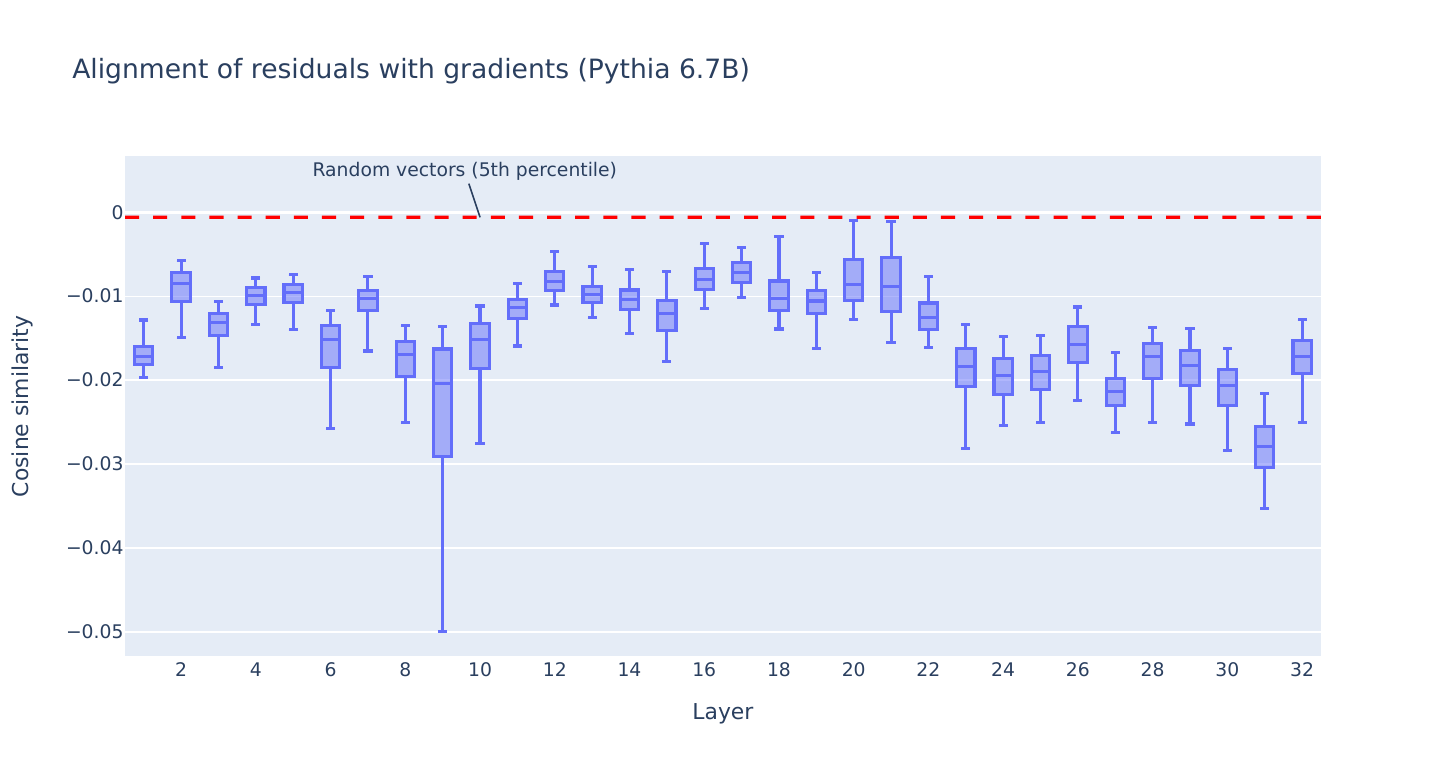}
    \includegraphics[scale=0.4, clip]{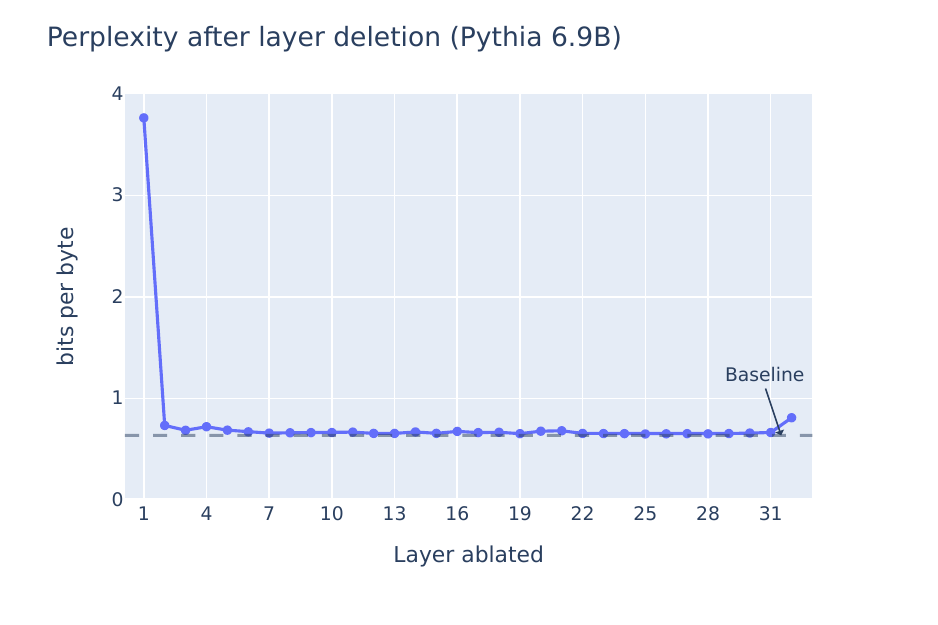}
    \caption{\textbf{Left:} Cosine similarity between the observed update to the hidden state and $ \frac{\partial \mathcal{L}(\hh_{i})}{\partial \hh_{i}}  $. The similarity is almost always negative, and of much larger magnitude than would be expected by chance in this very high dimensional space. Boxplot whiskers are 5th and 95th percentiles. Results were computed over a sample of roughly 16M tokens from the Pile validation set.
    \textbf{Right:} Cross-entropy loss of Pythia 6.9B on the Pile validation set, after replacing each of its 32 layers with the identity transformation. The dashed line indicates Pythia's perplexity on this dataset with all layers intact.}
    \label{fig:ablation-and-alignment}
\end{figure}

We reproduced their experiment using Pythia 6.9B, and report the results in Figure \ref{fig:ablation-and-alignment}. We show that, for every layer in the network, the cosine similarity between the residual and the gradient is negative at least 95\% of the time. While the magnitudes of the cosine similarities are relatively small in absolute terms, never exceeding 0.05, we show that they are much larger than would be expected of random vectors in this very high dimensional space. Specifically, we first sample 250 random Gaussian vectors of the same dimensionality as $ \frac{\partial \mathcal{L}(\hh_{i})}{\partial \hh_{i}}  $, which is (hidden size) $\times$ (sequence length) $ = 8{,}388{,}608$. We then compute pairwise cosine similarities between the vectors, and find the 5\textsuperscript{th} percentile of this sample to be $\mathbf{-6 \times 10^{-4}}$. Virtually all of the gradient-residual pairs we observed had cosine similarities below this number.

\subsection{Zero-shot robustness to layer deletion}
\label{app:layer-deletion}

Stochastic depth \citep{huang2016deep}, also known as LayerDrop \citep{fan2019reducing}, is a regularization technique which randomly drops layers during training, and can be applied to both ResNets and transformers. \citet{veit2016residual} show that ResNets are robust to the deletion of layers even when trained without stochastic depth, while CNN architectures without skip connections are not robust in this way. These results strongly support the idea that adjacent layers in a ResNet encode fundamentally similar representations \citep{greff2016highway}.

To the best of our knowledge, this experiment had never been replicated before in transformers. We did so and report our results in Figure \ref{fig:ablation-and-alignment} above. We find that only the very first layer is crucial for performance; every other layer ablation induces a nearly imperceptible increase in perplexity. Interestingly, \citet{veit2016residual} also found that the first layer is exceptionally important in ResNets, suggesting that this is a general phenomenon.

\section{Causal intervention details}

\begin{figure}
    \centering
    \includegraphics[scale=0.45]{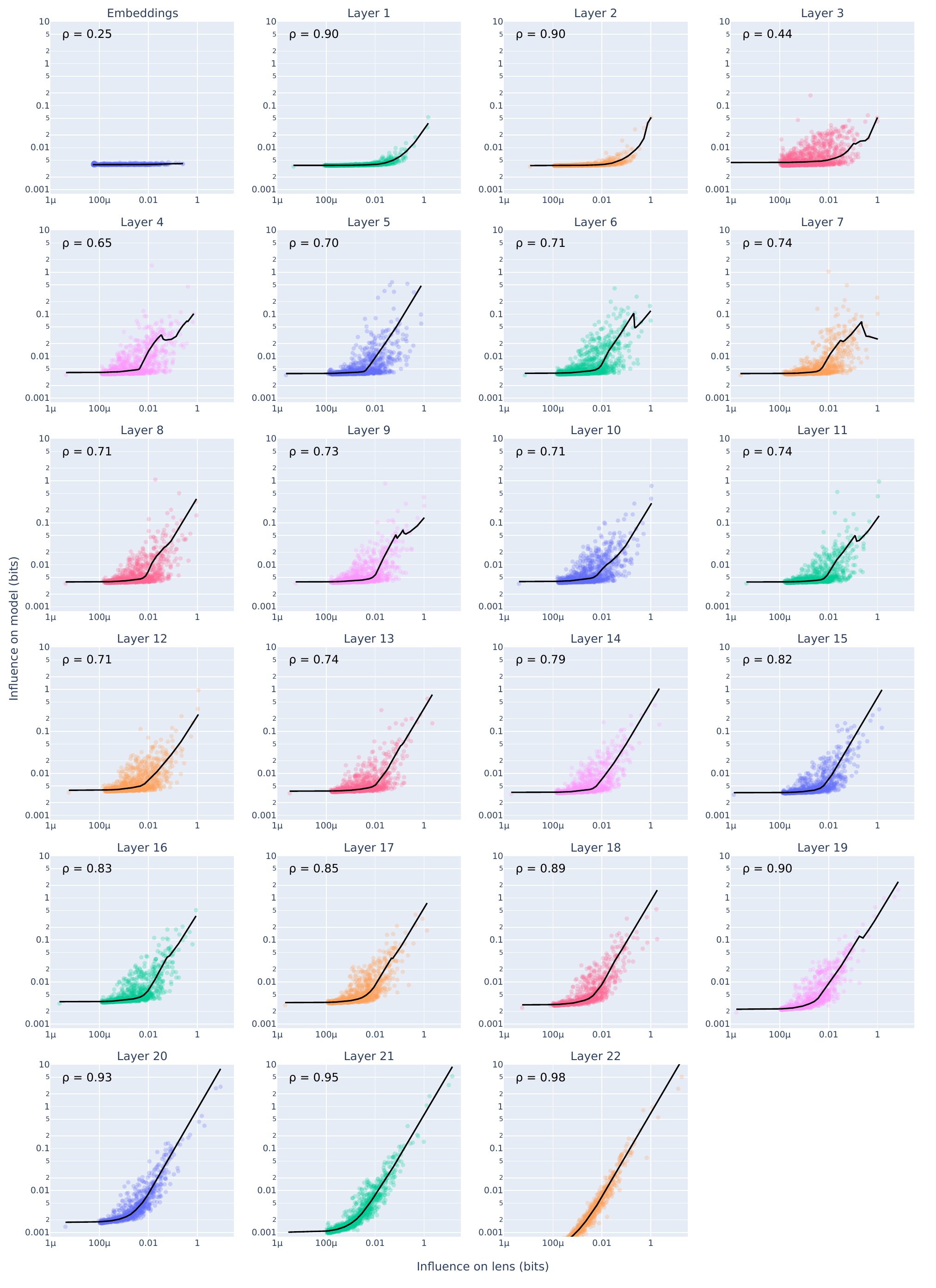}
    \caption{Causal fidelity in Pythia 410M across layers. In the top left corner of each plot is the Spearman rank correlation between the causal influence of each feature on the tuned lens and its influence on the model.}
    \label{fig:full-fidelity}
\end{figure}

\subsection{Ablation techniques}
\label{app:ablation-techniques}

Prior work has assumed that, given a representation $\xx$ and a linear subspace $B \subseteq \R^D$ along which we want to remove information, the best erasure strategy is to project $\xx$ onto the orthogonal complement of $B$:
\begin{equation}
    \xx' = {\projBComp(\xx)}
\end{equation}
This ``zeroes out'' a part of $\xx$, ensuring that $\innerproduct{\xx'}{\uu} = \boldsymbol 0$ for all $\uu$ in $B$. But recent work has pointed out that setting activations to zero may inadvertently push a network out of distribution, making the experimental results hard to interpret \citep{wang2022interpretability, chan2022causal}.

For example, a probe may rely on the invariant that $\forall \xx, \innerproduct{\xx}{\uu} \gg \boldsymbol 0$ for some $\uu$ in $B$ as an implicit bias term, even if $\innerproduct{\xx}{\uu}$ has low variance and does not encode any useful information about the input. The naive projection $\xx' = {\projBComp(\xx)}$ would significantly increase the loss, making it seem that $\projB{\xx}$ contains important information-- when in reality model performance would not be significantly degraded if $\projB{\xx}$ were replaced with a nonzero constant value more representative of the training distribution.

To correct for this, in our causality experiments we ablate directions by replacing them with their \emph{mean} values computed across a dataset, instead of zeroing them out. Specifically, to ablate a direction $\uu$, we use the formula:
\begin{equation}
    \xx' = \xx + P_{\uu}(\overline{\xx} - \xx)
\end{equation}
where $P_{u}$ is the projection matrix for $\uu$ and $\overline{\xx}$ is the mean representation.

\section{Static interpretability analysis}
\label{app:static-analysis}

Several interpretability methods aim to analyze parameters that ``write" (in the sense of \citet{elhage2021mathematical}) to intermediate hidden states of a language model \citep{millidge2022svd, dar2022analyzing, geva2022transformer, geva2020transformer}, and even use this analysis to successfully edit model behavior \citep{geva2022transformer, millidge2022svd, dai-etal-2022-knowledge}. Given that the tuned lens aims to provide a ``less biased" view of intermediate hidden states, we should expect the tuned lens to preserve their effectiveness. We test both static analysis and model editing methods, and confirm this hypothesis. With static analysis, the tuned lens appears to never decrease performance, and for some models, even increases their performance. With model editing, we found the tuned lens to outperform the logit lens on OPT-125m and perform equivalently on other models for the task of toxicity reduction.

\subsection{Static Analysis}

Many parameters in transformer language models have at least one dimension equal to that of their hidden state, allowing us to multiply by the unembedding matrix to project them into token space. Surprisingly, the resulting top $k$ tokens are often \textit{meaningful}, representing interpretable semantic and syntactic clusters, such as ``prepositions'' or ``countries'' \citep{millidge2022svd, dar2022analyzing, geva2022transformer}. This has successfully been applied to the columns of the MLP output matrices \citep{geva2022transformer}, the singular vectors of the MLP input and output matrices \citep{millidge2022svd}, and the singular vectors of the attention output-value matrices $W_{OV}$ \citep{millidge2022svd}.

In short\footnote{The exact interpretation depends on the method used - for instance, the rows of $W_{\text{out}}$ can be interpreted as the values in a key-value memory store \citep{geva2020transformer}.}, we can explain this as: many model parameters directly modify the model's hidden state \citep{elhage2021mathematical}, and when viewed from the right angle, the modifications they make are often interpretable. These ``interpretable vectors'' occur significantly more often than would be expected by random chance \citep{geva2022transformer}, and can be used to edit the model's behaviour, like reducing probability of specific tokens \citep{millidge2022svd}, decreasing toxicity \citep{geva2022transformer}, or editing factual knowledge \citep{dai-etal-2022-knowledge}.

Although these interpretability methods differ in precise details, we can give a generic model for the interpretation of a model parameter $W$ using the unembedding matrix:\footnote{In most transformer language models, the function here should technically be $T(W) = \text{top-}k(\text{softmax}(\text{LN}(f_i(W)W_U)))$, including both softmax and LayerNorm. However, the softmax is not necessary because it preserves rank order, and the LayerNorm can be omitted for similar reasons (and assuming that either $f_i(W)$ is zero-mean or that $W_U$ has been left-centered).}
$$T(W) = \text{top-}k(f_i(W)W_U) $$
where $f_i$ is some function from our parameter to a vector with the same dimensions as the hidden state, and $W_U$ is the unembedding matrix. In words: we extract a hidden state vector from our parameter $W$ according to some procedure, project into token space, and take the top $k$ matching tokens. The resulting $T(W)$ will be a list of tokens of size $k$, which functions as a human interpretable view of the vector $f_i(W)$, by giving the $k$ tokens most associated with that vector. As an example, the model parameter $W$ could be the MLP output matrix $W_{\text{out}}$ for a particular layer, and $f_i$ the function selecting the $i$th column of the matrix.

With the tuned lens, we modify the above to become:
$$T(W) = \text{top-}k(L_\ell(f_i(W)) W_U) $$
where $L_\ell$ is the tuned lens for layer number $\ell$, projecting from the hidden state at layer $\ell$ to the final hidden state.

To test this hypothesis, we developed a novel automated metric for evaluating the performance of these parameter interpretability methods, based on the pretrained BPEmb encoder \citep{heinzerling2018bpemb}, enabling much faster and more objective experimentation than possible with human evaluation. We describe this method in Appendix~\ref{app:static-analysis:automated}.

\begin{table*}[t]
\centering
\resizebox{\textwidth}{!}{\begin{tabular}{l|llllllll}
\hline
& OV SVD (L) & OV SVD (R) & QK SVD (L) & QK SVD (R) & $W_{in}$ SVD & $W_{in}$ columns & $W_{out}$ SVD & $W_{out}$ rows \\ \hline
Logit lens (real)   & 0.0745        & 0.1164         & 0.0745        & 0.1164         & 0.0745       & 0.0864           & 0.0728        & 0.0796         \\
Logit lens (random) & 0.0698        & 0.0695         & 0.0697        & 0.0688         & 0.0689       & 0.0689           & 0.0691        & 0.0688         \\
Tuned lens (real)   & 0.1240        & 0.1667         & 0.1240        & 0.1667         & 0.1193       & 0.1564           & 0.1196        & 0.1630         \\
Tuned lens (random) & 0.1164        & 0.1157         & 0.1177        & 0.1165         & 0.1163       & 0.1262           & 0.1163        & 0.1532         \\ \hline
\end{tabular}}
\vspace*{-.1in}
\caption{The mean interpretability scores (as measured in \cref{app:static-analysis:automated}) for Pythia 125M, with several different interpretability techniques \citep{millidge2022svd, geva2020transformer}, comparing both the tuned lens and logit lens to randomly generated matrices. Where applicable, the notation (L) and (R) indicates that the results are for the left and right singular vectors, respectively.}
\label{tab:interp-scores}
\end{table*}

\footnotetext{Random shuffling applied to each matrix (head-wise for attention matrices), to approximate the element-wise marginal distribution.}

The results for Pythia 125M can be seen in Table \ref{tab:interp-scores}. The parameters under both the tuned and logit lens consistently appear more interpretable than random. And the tuned lens appears to show benefit: the difference between random/real average scores is consistently higher with the tuned lens than the logit lens. However, preliminary experiments found much less improvement with larger models, where both the tuned and logit lens appeared to perform poorly.

\subsection{Interpretability score distributions}
\label{app:static-analysis:distributions}

\begin{figure}[t]
\hspace*{-.3in}
\includegraphics[width=3.7in]{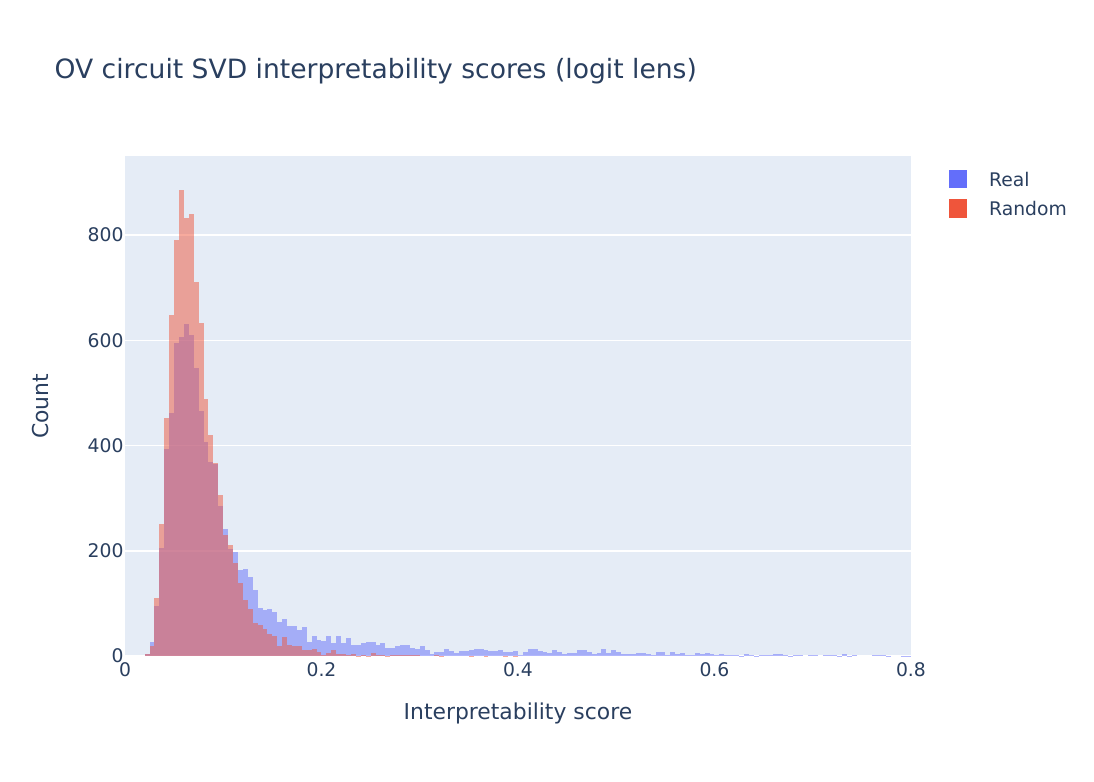}
\includegraphics[width=3.7in]{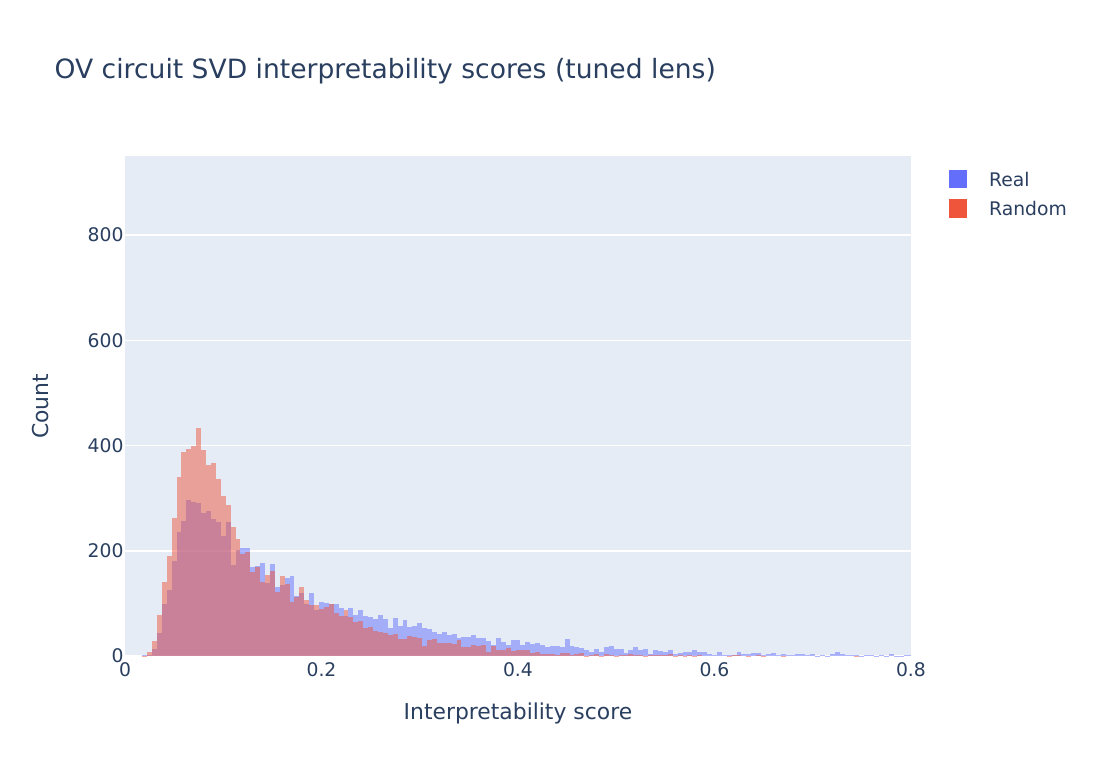}
\vspace*{-.4in}
\caption{The interpretability scores for the right singular vectors of the OV matrices $W_{OV}$ (following \citet{millidge2022svd}) in Pythia 125M, compared with a randomly generated matrix, for both the logit lens (left), and the tuned lens (right).}
\label{fig:interp-hist}
\end{figure}

The distribution of interpretability scores for a single parameter (the OV circuit) can be seen in Figure \ref{fig:interp-hist}. (Interpretability distributions of other parameters appear to share these features as well.) The plot displays a major complication for static analysis methods: most singular vectors are not more interpretable than random, except for a minority lying in a long right tail. It also shows the necessity of comparison to randomly generated matrices: naively comparing the interpretability scores for the tuned lens against those for the baseline would imply a far greater improvement than actual, because the interpretability scores under the tuned lens are higher even for randomly generated matrices.

\footnotetext{Similar to above footnote, random shuffling applied to each $W_{OV}$ for each head.}

\subsection{Automated interpretability analysis method}
\label{app:static-analysis:automated}

While humans can easily tell if a list of words represents some coherent category, this can be subjective and time-consuming to evaluate manually. Some previous work has attempted to automate this process, using GPT-3 in \citet{millidge2022svd} or the Perspective API in \citet{geva2022transformer}, but these can still be too slow for quick experimentation.

We created our own method, motivated by the observation that much of the human interpretable structure of these tokens appears to be related to their mutual similarity. Then, the mutual similarity of words can easily be measured by their cosine similarity under a word embedding \citep{Sitikhu2019ACO}. In particular, we use the pretrained BPEmb encoder \citep{heinzerling2018bpemb} \footnote{We also tried the fastText encoder \citep{bojanowski2016enriching}, but found BPEmb to give results better matching human judgment.}. We encode each token, measure the cosine similarity pairwise between all tokens, and take the average value:
$$I(T(W)) = \frac{\sum_{i,j}^{k,k} E(T_i(W)) \cdot E(T_j(W))}{k^2}$$
where the subscript on $T_i(W)$ denotes the $i$-th token of $T(W)$, and $E$ denotes the normalized BPEmb encoding function. This creates an interpretability metric $I$ that measures something like the ``monosemanticity" of a set of tokens.

Because BPEmb is a pretrained model, unlike most CBOW or skip-gram models, there is no ambiguity in the training data used to produce word similarities and results are easy to replicate. We find the interpretability scores given by the model to correspond with human judgment in most cases: it rarely identifies human-uninterpretable lists of tokens as interpretable, although it occasionally struggles to recognize some cases of interpretable structure, like syntactically related tokens, or cases where a list as a whole is more interpretable than a pairwise measure can capture.

\subsection{Improved Model Editing}

\citet{geva2022transformer} edit GPT-2-medium to be less toxic by changing the coefficients which feed into the columns of the MLP output matrix, under the key-value framework developed in \citet{geva2020transformer}. They define both the value vector $v_i^\ell$ (i.e. the columns of the MLP's $W_{\text{out}}$) and a coefficient $m_i^\ell$ (i.e. the output of $W_{\text{in}}$ before the activation function) for each column $i$ and each layer $\ell$. This is used to reduce the toxicity of GPT-2-medium by editing the coefficients of specific value vectors. They find:
$$T(W_{\text{out}^l}) = \text{top-}30(f_i(W_{\text{out}}^\ell)W_U) $$
where: 
$$f_i(W_{\text{out}}^\ell) = v_i^\ell$$

They then concatenate these top-30 tokens for every layer and column, sending it to Perspective API, a Google API service which can return a toxicity score. They then sampled from the vectors that scored < 0.1 toxicity (where a score > 0.5 is classified as toxic), set each value vector's respective coefficient to 3 (because 3 was a higher than average activation), and ran the newly edited model through a subset of REALTOXICPROMPTS \citep{gehman-etal-2020-realtoxicityprompts} to measure the change in toxicity compared to the original model. 

Changing the coefficients of value vectors can generate degenerate text such as always outputting `` the" which is scored as non-toxic. To measure this effect, they calculate the perplexity of the edited models, where a higher perplexity may imply these degenerate solutions. 

We perform a similar experiment, but found the vast majority of value vectors that scored < 0.1 toxicity to be non-toxic as opposed to anti-toxic (e.g. semantic vectors like `` the", `` and", etc are scored as non-toxic). Alternatively, we set the $k$ most toxic value vector's coefficient to 0, as well as projecting with both the logit and tuned lens on OPT-125m, as seen in Table \ref{table:toxicity-scores}. 

We additionally tested pythia-125m, pythia-350m, gpt-2-medium, and gpt-neo-125m; however, there were no significant decreases in toxicity (i.e. > 2\%) compared to the original, unedited models.

\begin{table*}[t]
\centering
\begin{tabular}{l|lllllll}
\hline
  & Toxicity & \begin{tabular}[c]{@{}l@{}}Severe\\ Toxicity\end{tabular} & \begin{tabular}[c]{@{}l@{}}Sexually\\ explicit\end{tabular} & Threat & Profanity & \begin{tabular}[c]{@{}l@{}}Identity\\ attack\end{tabular} & Perplexity \\ \hline
Original/ Unedited         & 0.50     & 0.088                                                     & 0.159                                                       & 0.058  & 0.40      & 0.043                                                     & \textbf{27.66}      \\
Logit Lens Top-5  & 0.47     & 0.077                                                     & 0.148                                                       & 0.057  & 0.38      & 0.043                                                     & 27.68      \\
Logit Lens Top-10 & 0.45     & 0.073                                                     & 0.144                                                       & 0.056  & 0.36      & 0.040                                                     & 27.70      \\
Logit Lens Top-20 & 0.43     & 0.072                                                     & 0.141                                                       & 0.057  & 0.33      & 0.041                                                     & 27.73      \\
Tuned Lens Top-5  & 0.45     & 0.079                                                     & 0.143                                                       & 0.056  & 0.36      & 0.041                                                     & \textbf{27.66}      \\
Tuned Lens Top-10 & 0.42     & 0.063                                                     & 0.143                                                       & 0.057  & 0.33      & \textbf{0.034}                                                     & 27.67      \\
Tuned Lens Top-20 & \textbf{0.39}     & \textbf{0.061}                                                     & \textbf{0.138}                                                       & \textbf{0.052}  & \textbf{0.31}      & 0.035                                                     & 27.73      \\ \hline
\end{tabular}
\caption{Percentage of prompt generations labeled as toxic for OPT-125m. Best results in \textbf{bold}. Across all respective settings, the tuned lens outperformed the logit lens with no significant increases in perplexity. Although the value vectors were only graded \& ranked by toxicity, there were still decreases in other metrics (severe toxicity, sexually explicit, etc) that were not directly selected against. However, these results did not generalize to other models tested.}
\label{table:toxicity-scores}
\end{table*}

\subsection{Implementation}

 The initial implementation of many of our static analysis experiments was helped by the use of the \texttt{transformer\_lens} library \citep{nandatransformerlens2022}.


\end{document}